\documentclass[
 10pt,
 onecolumn,
 superscriptaddress,
 amsmath,amssymb,
 aps,
 pra,
 showkeys,
 floatfix,
]{revtex4-2}

\usepackage{graphicx}
\usepackage{dcolumn}
\usepackage{bm}
\usepackage{xcolor}
\usepackage{booktabs}
\usepackage{makecell}
\usepackage{array}
\usepackage[normalem]{ulem} 
\usepackage{siunitx}
\usepackage{algorithmicx}
\usepackage{algpseudocode}
\usepackage{pifont}
\newcommand{\best}[1]{\textbf{#1}}
\newcommand{\second}[1]{\underline{#1}}
\newcommand{\cmark}{\ding{51}} 
\usepackage{hyperref}

\begin{document}

\title{Periodic Topological Deep Learning for Polymer Design and Discovery}
\author{Yasharth Yadav}
\thanks{To whom correspondence should be addressed: yasharth001@e.ntu.edu.sg or xiakelin@ntu.edu.sg}
\affiliation{School of Physical and Mathematical Sciences, Nanyang Technological University, Singapore 637371}

\author{Tze Kwang Gerald Er}
\affiliation{School of Chemistry, Chemical Engineering and Biotechnology, Nanyang Technological University, Singapore 637371}

\author{Atsushi Goto}
\affiliation{School of Chemistry, Chemical Engineering and Biotechnology, Nanyang Technological University, Singapore 637371}

\author{Kelin Xia}
\thanks{To whom correspondence should be addressed: yasharth001@e.ntu.edu.sg or xiakelin@ntu.edu.sg}
\affiliation{School of Physical and Mathematical Sciences, Nanyang Technological University, Singapore 637371}

\begin{abstract}
Polymers underpin applications across energy, healthcare, and materials science, yet their vast chemical space makes systematic discovery challenging. Most machine learning approaches represent polymers as molecular graphs of a single repeating unit, thereby missing both the periodicity of polymer chains and many-body interactions beyond pairwise bonds. We introduce Periodic-TDL, a deep learning framework built on periodic Rips complexes that capture many-body interactions across multiple spatial scales, followed by a hierarchical simplicial message-passing (HSMP) encoder that propagates information from long-range interactions to covalent bonds, yielding representations enriched by higher-order topological features. Periodic-TDL achieves state-of-the-art performance across polymer property prediction tasks spanning electronic, optical, physical, and thermal targets. Furthermore, we use Periodic-TDL to quantitatively validate how ester-to-amide substitution and $\alpha$-methylation enhance thermal stability. Applying the model to a computationally synthesized dataset of 48208 structures generated via systematic substitution of acrylate and acrylamide polymers, we observed a mean $T_g$ increase of $\sim 55^\circ$C for ester-to-amide substitutions and $\sim 14^\circ$C for backbone $\alpha$-methylation across matched polymer pairs. Finally, we use Periodic-TDL to analyze six novel polymer pairs from independent experimental measurements, including three newly synthesized polymers previously unreported in the literature. The experimental data successfully confirmed the trends predicted by Periodic-TDL. Overall, our results demonstrate that a periodic topological representation of polymers effectively enhances predictive performance and captures the physical effects of functional group modifications.
\end{abstract}

\maketitle
\section*{Introduction}\label{sec:Intro}
Polymers underpin modern life across electronics, aerospace, energy, and healthcare \cite{geyer2017production, lopez2019designing, leigh2020helical}, owing to their tunable thermal, electronic, mechanical, and optical properties. 
This tunability arises from a virtually infinite chemical space that can be navigated by modifying their molecular architecture and functional group chemistry \cite{mccrum1997principles, coleman2019fundamentals}. 
While the intuition of polymer chemists remains invaluable, the sheer scale of this space renders purely experimental discovery impractical \cite{feldman2008polymer}. 
Polymer informatics has emerged to address this challenge by applying modern tools from data science and machine learning, enabling substantial progress in both property prediction and inverse design, where polymers meeting prescribed requirements are identified \cite{audus2017polymer, chen2021polymer, wu2019machine, barnett2020designing, kuenneth2022bioplastic}. 

Central to any machine learning approach is the choice of how polymer structure is encoded. 
Early methods derived hand-crafted numerical vectors from local atomic environments, functional groups, and packing descriptors \cite{rogers2010extended, kim2018polymer, doan2020machine, axelrod2022learning}. 
With the rise of deep learning, the field shifted toward learning directly from machine-readable molecular representations, where numerical vectors emerge as an implicit outcome of the neural encoder. 
This shift places the burden of accuracy squarely on the representation itself, which must provide a faithful and complete description of the molecular structure for the encoder to extract meaningful information. 
Sequence-based Transformers \cite{xu_transpolymer_2023, kuenneth_polybert_2023, alagarsamy2026transformers} treat the SMILES string \cite{weininger1988smiles} as a chemical language, pretraining on large unlabelled corpora before fine-tuning on labelled property datasets. 
Message-passing neural networks \cite{chen2019graph, st2019message, wang_molecular_2022, mohapatra2022chemistry, antoniuk_representing_2022, aldeghi_graph_2022, heid2023chemprop, queen_polymer_2023, gurnani_polymer_2023} learn from molecular graphs describing covalent connectivity. 
Multimodal frameworks \cite{qiu2024polync, wang_mmpolymer_2024, parambil_transchem_2025, huang2025unified} combine different representation types to compensate for the blind spots of any single modality. 
Despite this diversity, most deep learning frameworks construct their structural input from a single, finite repeating unit. 
As a result, they effectively treat polymers as small molecules rather than as macromolecules, creating a fundamental representational bottleneck that remains largely unaddressed \cite{gao2024machine, gao2026ai, liu2025open}.

Unlike small molecules with a fixed number of atoms, polymers consist of periodic chains of repeating chemical subunits that can extend to millions of atoms and adopt complex architectures \cite{yamauchi2021two}. A faithful representation must therefore encode structural relationships that span adjacent repeating units. Several works have proposed periodic graph representations as a more suitable alternative. The wDMPNN framework \cite{aldeghi_graph_2022}, the periodic polymer graph \cite{antoniuk_representing_2022}, and PolyGNN \cite{gurnani_polymer_2023} extend beyond an isolated repeating unit to capture inter-unit connectivity under periodic boundary conditions. More recently, PerioGT \cite{wu2025periodicity} introduced periodicity augmentation and contrastive pre-training to align representations of equivalent polymer fragments, along with graph-transformer attention and periodicity prompts during fine-tuning. However, these approaches remain fundamentally graph-based, with structural interactions expressed through pairwise relations between atoms. Particularly, higher-order relationships involving three or more atoms are not represented explicitly. These limitations call for a richer representational framework that directly encodes polymer periodicity while explicitly encoding many-body interactions across multiple spatial scales.

More recently, topological deep learning (TDL) has emerged as a principled extension of graph neural networks that addresses their fundamental restriction to pairwise interactions \cite{hajij2022topological, hajij2023combinatorial, papamarkou2024position, pham2025topological}. 
Rather than representing molecular structure as a graph of atoms and bonds, TDL operates on richer topological domains that can natively encode many-body interactions among three or more elements simultaneously. 
The most widely studied of these are simplicial complexes, which are generalizations of graphs capable of encoding interactions between multiple atoms through higher-dimensional simplices \cite{carlsson_topology_2009, edelsbrunner_computational_2010, otter_roadmap_2017}. 
Other structures, including cell complexes \cite{forman_bochners_2003}, sheaves \cite{hansen2019toward}, and hypergraphs \cite{bick2023higher}, are also gaining traction. 
Message passing schemes defined across these structures enable information exchange between objects of varying dimension. 
This allows representations to incorporate both local and global structural information \cite{bodnar_weisfeiler_2021, bodnar2021weisfeiler, schaub2022signal, giusti2023cell}. 
These capabilities have demonstrated considerable success in molecular property prediction \cite{cang2017topologynet, wee2025review, giusti2024topological, amarjeet2026topological, shen2026molecular}. 
However, no existing TDL framework incorporates the periodic nature of polymers, where chains are still treated as finite molecular fragments, discarding the key structural aspect that distinguishes polymers from small molecules.

Here we introduce Periodic-TDL, a framework that captures both periodicity and higher-order interactions across multiple spatial scales. 
At its core is the \textit{periodic Vietoris–Rips complex} representation of polymers, which encodes interactions between atoms across adjacent repeating units under periodic boundary conditions, representing both covalent and non-covalent interactions through higher-dimensional simplices. 
To learn from this representation, we develop the \textit{Hierarchical Simplicial Message Passing (HSMP)} encoder, which operates over a nested sequence of simplicial complexes and propagates information from coarser to finer scales in a chemically motivated direction. 
Notably, at the finest filtration level, HSMP reduces to a standard molecular graph, enabling direct integration into established self-supervised pretraining pipelines \cite{rong_self-supervised_2020, wang_molecular_2022, zang2023hierarchical, wang2023evaluating, zhang2021motif, koo2025comprehensive} designed for conventional graph encoders. 
This compatibility is particularly valuable given the scarcity of labelled polymer data. 
To equip higher-order simplices with meaningful features, we further introduce curvature-based featurization using Forman's discretization of Ricci curvature \cite{forman_bochners_2003}, which generalizes naturally to simplices of any dimension. 
Next, we pretrained Periodic-TDL on one million unlabelled polymers and evaluated its performance across nine polymer properties spanning electronic, optical, physical, and thermal targets.
Finally, we applied Periodic-TDL to a computationally generated dataset of 48208 systematically substituted polymers from acrylate and acrylamide families. 
Specifically, we studied the effects of ester-to-amide substitution and backbone $\alpha$-methylation on the glass transition temperature ($T_g$) of these polymers.
To ensure the chemical credibility of our model, we compared the predicted $T_g$ changes against six polymer pairs spanning eight polymers from prior experimental studies and three newly synthesized polymers.


\section*{Results}\label{sec:Results}
\subsection*{Periodic Rips Filtration Captures Topology Beyond a Single Repeating Unit}
We represent linear homopolymers using a periodic Vietoris–Rips complex that formally encodes topological relationships between atoms belonging to adjacent repeating units. 
Technical details of our construction are provided in the Methods section. Here, we illustrate the core ideas using poly(bisphenol A carbonate) as a representative example.

\begin{figure}[htp]
    \centering
    \includegraphics[width=0.8\linewidth]{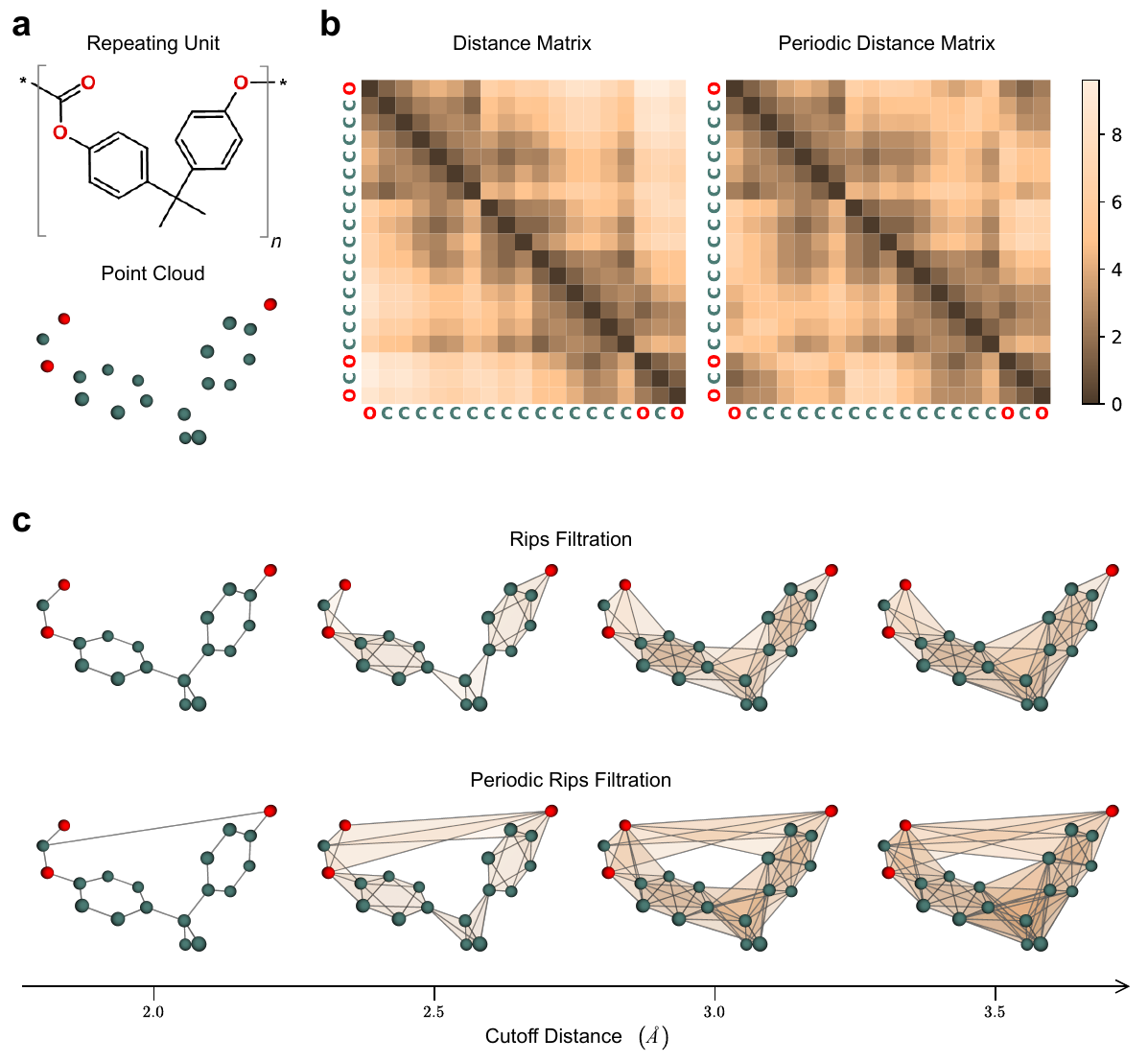}
    \caption{ \textbf{Periodic Vietoris–Rips complex representation of poly (bisphenol A carbonate).}
    (a) Repeating unit of poly (bisphenol A carbonate) and its corresponding three-dimensional atomic point cloud. 
    (b) Pairwise distance matrices computed from a single repeating unit (left, intra-monomer distance matrix) and from its translated copes (right, periodic distance matrix). The periodic distance matrix restores the correct spatial proximity between atoms that lie near the boundaries of a chosen repeating unit but are adjacent in the actual polymer. 
    (c) Comparison of Vietoris–Rips filtration constructed using the intra-monomer distance matrix (top row) and the periodic distance matrix (bottom row) at increasing cutoff distances. }
    \label{fig:PerRepFig}
\end{figure}

Figure \ref{fig:PerRepFig}(a) shows the repeating unit of poly (bisphenol A carbonate)  and its corresponding three-dimensional atomic point cloud. 
A pairwise Euclidean distance matrix constructed from this point cloud [Figure \ref{fig:PerRepFig}(b)] encodes spatial proximity only within a single arbitrarily chosen unit, failing to account for translational symmetry along the polymer backbone. 
This limitation is particularly evident at the boundaries, where atoms belonging to the carbonate linkage appear spatially distant simply because they reside at opposite ends of the chosen unit, despite being directly connected in the actual polymer. 
To resolve this, we construct a periodic distance matrix by incorporating translated copies of the repeating unit via iterative rearrangement of backbone fragments while strictly preserving chemical connectivity \cite{lo_augmenting_2023}, defining each interatomic distance as the minimum separation across all translated copies.

From this periodic distance matrix, multi-scale topological structure is extracted via Vietoris-Rips filtration \cite{carlsson_topology_2009, edelsbrunner_computational_2010}. 
For a given cutoff $\epsilon_i$, any set of $k+1$ atoms with all pairwise distances below $\epsilon_i$ gives rise to a $k$-simplex, and the resulting simplicial complex $\mathrm{VR}_{\epsilon_i}(\mathbf{D})$ acts as a higher-dimensional generalization of a graph. 
Progressively increasing $\epsilon_i$ incorporates interactions over increasing spatial ranges, capturing both covalent and non-covalent interactions \cite{cang_representability_2018}. 
Simplicial complexes and Vietoris–Rips filtration are formally defined in Supplementary Material.

Figure \ref{fig:PerRepFig}(c) compares Vietoris–Rips filtration constructed using the non-periodic (intra-monomer) distance matrix and the periodic distance matrix. 
In both cases the same filtration procedure is applied, but the underlying distance matrices differ, leading to distinct simplicial complexes. 
In the non-periodic representation, atoms located near the boundaries of the chosen repeating unit appear spatially distant and therefore remain disconnected at small cutoffs. 
In contrast, the periodic distance matrix restores the correct spatial proximity between atoms belonging to adjacent repeating units, yielding a Vietoris–Rips complex that reflects the connectivity of the extended polymer and is invariant to the arbitrary choice of repeating unit. 
At $\epsilon_i=2 \text{\AA}$, the resulting complexes recover covalent bonds. 
In the non-periodic case, these correspond only to bonds within a single repeating unit, whereas the periodic Vietoris–Rips complex captures covalent bonds across repeating units along the polymer backbone. 
At cutoffs of $\epsilon_i=2.5 \text{\AA}$ and beyond, higher-dimensional simplices begin to appear, reflecting many-body interactions. 
These simplicial complexes grow denser at larger cutoffs as more atoms fall within mutual proximity.  
Thus, the periodic Vietoris–Rips filtration accurately represents the topological structure of a polymer chain while systematically incorporating multi-body interactions across varying spatial scales

\subsection*{HSMP Learns Across Simplex Dimensions and Filtration Scales}
We develop the \textit{hierarchical simplicial message passing (HSMP) encoder} as a  learning component of Periodic-TDL, designed to operate directly on the periodic Vietoris–Rips complex constructed above. Since a Vietoris–Rips complex explicitly includes simplices of multiple dimensions arising from a filtration across varying spatial scales, the HSMP encoder operates both within and across individual filtration levels, enabling the integration of 
multi-scale interactions in a principled manner.

\subsubsection*{Multi-head simplicial message passing}
We extend the Message Passing Simplicial Network framework of Bodnar et al. \cite{bodnar_weisfeiler_2021} to a multi-head setting, allowing multiple parallel simplex updates that operate on different representation subspaces to capture complementary 
interaction patterns \cite{bodnar_weisfeiler_2021, goh2022simplicial, yang_simplicial_2022}. At a given filtration parameter, let $\sigma \in \mathrm{VR}_{\epsilon_i}(\mathbf{D})$ be a simplex with feature $\mathbf{x}_{\sigma}$. As illustrated in Figure~\ref{fig:ArchFig}(a), $\mathbf{x}_{\sigma}$ is projected and split into $k$ parallel representations $\mathbf{h}_{\sigma}^{0,(i)} \in \mathbb{R}^{D/k}$, each processed independently through a $T$-layer message passing module and concatenated to yield the updated representation $\mathbf{z}_{\sigma}$.

Each message passing layer aggregates over the upper-adjacent neighborhood of $\sigma$, defined via the boundary incidence relation $\prec$, where $\sigma \prec \tau$ if $\sigma \subset \tau$ and there exists no $\delta$ such that $\sigma \subset \delta \subset \tau$. The co-boundary and upper-adjacent sets of $\sigma$ are respectively
\begin{align*}
\mathcal{C}_{\sigma} &= \{ \tau \mid \sigma \prec \tau \}, \\
\mathcal{N}^{\uparrow}_{\sigma} &=
\left\{
\sigma' \;\middle|\;
\exists\, \tau \in \mathcal{C}_{\sigma} \text{ such that } \sigma, \sigma' \prec \tau
\right\}.
\end{align*}
In other words, two simplices are upper-adjacent if they share a common coface. Simplex representations are updated as
$$
\mathbf{h}_{\sigma}^{t+1,(i)} 
=
\phi\!\left(
\mathbf{h}_{\sigma}^{t,(i)},
\underset{\sigma' \in \mathcal{N}^{\uparrow}_{\sigma}}{\mathrm{AGG}}
\;\psi\!\left(\mathbf{h}_{\tau}^{(i)}, \mathbf{h}_{\sigma'}^{t,(i)}\right)
\right),
$$
where $\tau$ is the shared coface of $\{\sigma, \sigma'\}$. The coface feature $\mathbf{h}_{\tau}^{(i)}$ contributes to the update of lower-dimensional simplices, analogous to generalized graph convolution operators \cite{li_deepergcn_2020}. Here,  $\mathrm{AGG}$ is softmax aggregation, $\psi$ is addition followed by $\mathrm{ReLU}$, and $\phi$ is addition followed by an $\mathrm{MLP}$.

\subsubsection*{Cross-scale refinement}
\textit{Cross-scale refinement} enriches finer-scale simplex representations using long-range interaction information (Figure~\ref{fig:ArchFig}(b)). Since any filtration is a nested sequence of subcomplexes, any simplex $\sigma \in \mathrm{VR}_{\epsilon_i}$ also exists in $\mathrm{VR}_{\epsilon_{i+1}}$, and we define its refined representation as
$$
\mathbf{x}'_{\sigma,\epsilon_i}
=
\mathbf{x}_{\sigma,\epsilon_i}
+
\mathbf{x}_{\sigma,\epsilon_{i+1}}
\odot
\mathrm{MLP}\!\Big(
\mathbf{x}_{\sigma,\epsilon_{i+1}}
\;\|\;
\mathrm{MLP}(\mathbf{x}_{\sigma,\epsilon_{i+1}})
\Big),
$$
where the coarse-scale representation $\mathbf{x}_{\sigma,\epsilon_{i+1}}$ is used to produce a gated modulation vector via $\mathrm{MLP}$, which is then added residually to the finer-scale features $\mathbf{x}_{\sigma,\epsilon_i}$ before the next round of multi-head message passing.

\subsubsection*{End-to-end pipeline}
The HSMP encoder combines the above two simplex update functions into a systematic pipeline for learning polymer representations, with a schematic provided in Figure~\ref{fig:ArchFig}(c). 
Starting from a pSMILES representation, we computed a periodic distance matrix and applied a Vietoris–Rips filtration at three cutoff distances, $\epsilon_3 = 4.0\,\text{\AA}$,  $\epsilon_2 = 3.0\,\text{\AA}$, and $\epsilon_1 = 2.0\,\text{\AA}$. 
Initial simplex features comprise atom and bond descriptors commonly used in molecular GNNs \cite{rittig_graph_2023, gurnani_polymer_2023}, augmented with Forman's discretization of Ricci curvature \cite{forman_bochners_2003, samal_comparative_2018, sreejith_forman_2016, sreejith_systematic_2017} as a geometry-aware simplex feature applicable to any dimension \cite{wee2021forman, lai_deeper_2023, fesser2024effective}. 
Detailed feature descriptions and the formal definition of Forman curvature are provided in the Methods section and Supplementary Material.

Message passing proceeds hierarchically from the coarsest to the finest scale. 
At $\epsilon_3$, node and edge representations are updated via co-boundary and upper-adjacent relations, after which cross-scale refinement transfers information to $\epsilon_2$. 
This process repeats through to $\epsilon_1$, where message passing is restricted to the covalent bond graph. 
Here, atom and bond features are augmented by curvature descriptors from $\mathrm{VR}_{\epsilon_1}(\mathbf{D})$. 
This architecture enforces hierarchy along two directions. 
Within each filtration level, information flows from higher-order to lower-order structures. 
Across levels, it flows in order of decreasing cutoff. 
Consequently, atom and bond representations at the covalent scale already incorporate geometry-enhanced signals from higher-dimensional simplices and coarser filtration levels.

\subsubsection*{Pretraining and fine-tuning}
\begin{figure}[htp!]
    \centering
    \includegraphics[width=0.8\linewidth]{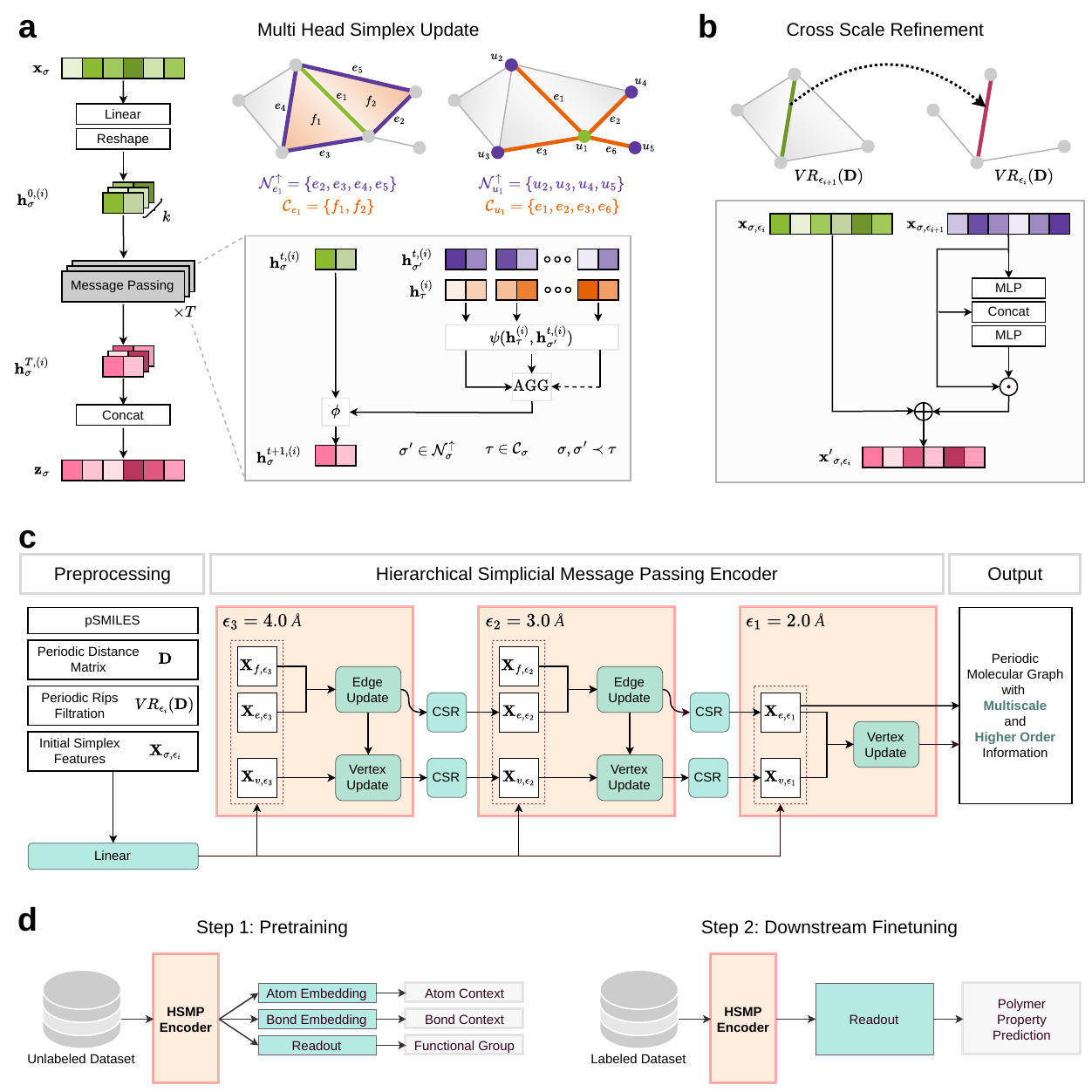}
    \caption{\textbf{Overview of hierarchical simplicial message passing.}
    (a) Multi-head simplicial message passing at a fixed filtration level $\mathrm{VR}_{\epsilon_i}(\mathbf{D})$, where simplex features are projected, split into multiple heads, updated via upper-adjacency (coface-mediated) message passing, and concatenated.
    (b) Cross-scale refinement between consecutive filtration levels $\mathrm{VR}_{\epsilon_{i+1}} \rightarrow \mathrm{VR}_{\epsilon_i}$, where coarse-scale simplex representations modulate finer-scale features through a residual gated mechanism.
    (c) The full HSMP pipeline operating over the Vietoris–Rips filtration, combining within-level simplicial message passing and cross-scale refinement across multiple cutoffs, proceeding from coarser to finer spatial scales.
    (d) Pretraining and fine-tuning pipelines, where the encoder is pretrained using atom context, bond context, and functional group prediction tasks at the finest level $\epsilon_1$, and finetuned by global mean pooling of atom embeddings for downstream polymer property prediction.}
    \label{fig:ArchFig}
\end{figure}

We pretrained HSMP using three self-supervised tasks adapted from the GROVER framework \cite{rong_self-supervised_2020}, namely atom context prediction, bond context prediction, and functional group (FG) prediction. 
The first two were formulated as node and edge classification tasks, where the model predicted discrete context labels encoding local chemical environments from atom and bond features. 
FG prediction is a polymer-level multi-label classification task requiring a global representation, obtained via mean pooling over atom embeddings:
$$
\mathbf{z}_{\mathrm{polymer}}
=
\frac{1}{|\mathcal{V}|}
\sum_{u \in \mathcal{V}} \mathbf{z}_u,
$$
where $\mathcal{V}$ denotes the set of atoms and $\mathbf{z}_u$ are the learned atom embeddings at $\epsilon_1$. This pooling operation was also applied during fine-tuning for downstream polymer property prediction, where pretrained weights provide an informed initialization that transfers both local and global chemical context. 
The pretraining and fine-tuning pipelines are illustrated in Figure~\ref{fig:ArchFig}(d).

\subsection*{Periodic-TDL Achieves Competitive Performance Across Diverse Polymer Properties}

We pretrained the HSMP encoder on one million polymers from the PI1M dataset \cite{ma_pi1m_2020} and finetuned it on nine downstream tasks spanning electronic ($E_{ea}$, $E_i$ \cite{kuenneth_polymer_2021}, $E_{gb}$, $E_{ib}$ \cite{kamal_novel_2021}, and $E_{gc}$), optical (EPS and $N_c$ \cite{kuenneth_polymer_2021}), physical ($X_c$ \cite{kuenneth_polymer_2021}), and thermal ($T_g$ \cite{malashin_estimation_2024}) properties. Here, $T_g$ contains experimentally measured values, whereas the remaining properties were derived from density functional theory calculations. Performance was assessed using five train/validation/test folds and reported as root-mean-square error (RMSE) and coefficient of determination ($R^2$) on the held-out test sets. Summary statistics for the nine datasets are provided in \textbf{Table~S1} and \textbf{Figure~S1}, and the experimental settings are described in Methods. 

The comparison included fingerprint-based Morgan neural networks \cite{morgan_generation_1965}, sequence-based polyBERT \cite{kuenneth_polybert_2023} and TransPolymer \cite{xu_transpolymer_2023}, graph-based polyGNN \cite{gurnani_polymer_2023}, MolCLR (GCN), and MolCLR (GIN) \cite{wang_molecular_2022}, multimodal TransChem \cite{parambil_transchem_2025} and MMPolymer \cite{wang_mmpolymer_2024}, and the periodicity-aware graph-transformer model PerioGT \cite{wu2025periodicity}. All models were evaluated using identical data splits. Baseline implementation details are provided in the Supplementary Material. The RMSE results are presented in \textbf{Table~\ref{tab:rmse_comparison}}, with the corresponding $R^2$ results reported in \textbf{Table~S2}.

Periodic-TDL achieved the lowest test RMSE on seven of the nine prediction tasks, namely $E_{ea}$, $E_i$, EPS, $N_c$, $X_c$, $E_{gb}$, and $E_{ib}$ (\textbf{Table~\ref{tab:rmse_comparison}}). The strongest baseline was PerioGT, a recently proposed pretrained graph transformer that encodes polymer periodicity through contrastive learning over chemically equivalent fragments sampled from different arrangements of the polymer repeat unit \cite{wu2025periodicity}. PerioGT yielded slightly lower errors than Periodic-TDL for $E_{gc}$ and $T_g$ datasets. The comparatively weaker performance of Periodic-TDL on these two targets may partly arise from differences in the information supplied to the models. Periodic-TDL learns directly from periodic molecular geometry and higher-order topological interactions, whereas PerioGT incorporates global chemical features derived from Morgan fingerprints and Mordred descriptors. These predefined features provide complementary information about functional groups and global physicochemical characteristics that may not be fully captured by the learned simplicial representation.

\subsubsection*{Including Additional Chemical Information and Residual Correction}
We evaluated two additional Periodic-TDL configurations that incorporate complementary chemical features. Periodic-TDL (+Chem) combines the fine-tuned Periodic-TDL prediction with predictors trained on frozen Periodic-TDL representations, Mordred descriptors, and Morgan fingerprints via a linear model. Periodic-TDL (+Chem, +RC) further applies a residual correction based on polymer similarity, using neighborhoods defined by Periodic-TDL embeddings and Morgan fingerprint similarity to estimate local prediction errors \cite{yang2023resmem}. Full details are provided in Methods. 

As shown in \textbf{Table~\ref{tab:rmse_comparison}}, incorporating global chemical descriptors selectively improved performance on several prediction tasks. Periodic-TDL (+Chem) achieved the largest gains for $E_{ib}$ and $T_g$, reducing the RMSE of the base Periodic-TDL model by 2.9\% and 4.1\%, respectively. Incorporating residual correction further improved performance on $E_{gc}$, $E_{gb}$, and $T_g$, with RMSE reductions of 1.4\%, 0.9\%, and 6.5\%, respectively, relative to the base model. These enhancements enabled Periodic-TDL (+Chem, +RC) to outperform PerioGT on $E_{gc}$ and $T_g$, where the original Periodic-TDL model was slightly less generalizable. By contrast, the original Periodic-TDL model remained the best-performing for $E_{ea}$, $E_i$, EPS, $N_c$, and $X_c$, indicating that descriptor-based predictors and local residual estimates do not consistently provide additional benefit when the periodic simplicial representation already captures the relevant structural information. The limited contribution of residual correction on these tasks is also consistent with the greater data requirements of neighborhood-based methods \cite{yang2023resmem}.

\subsubsection*{Ablation Experiments}
Ablation experiments independently examined the contributions of the principal Periodic-TDL design choices. Removing either the periodic representation or hierarchical message passing degraded predictive performance, demonstrating that both the treatment of the polymer as an extended periodic system and the propagation of information across simplex dimensions and filtration scales contribute to the model's predictive performance. Additional experiments varying multi-head simplicial message passing, self-supervised pretraining, and model capacity further showed that the complete pretrained multi-head configuration provided the strongest overall performance. Detailed ablation settings and results are reported in the Supplementary Material (\textbf{Tables~S3--S5}).

\subsection*{Predicted \texorpdfstring{$T_g$}{Tg} Trends in Acrylates and Acrylamides Are Chemically Credible}
Next, we evaluated whether the Periodic-TDL architecture captures two established trends in the glass transition temperature of acrylate- and acrylamide-based polymers, namely the effect of replacing ester with amide functional groups and the effect of backbone $\alpha$-methyl substitution. These trends are chosen because they are extensively documented in the experimental literature \cite{gallardo_effect_1993, aydogan_ayaz_influence_2013, fleischhaker_glass-transition-_2014, ruiz-rubio_influence_2015, zhou_high_2016, adharis_synthesis_2018, chuang_mechanical_2026, yuan_multifunctional_2026}. 
We evaluated chemical credibility in three steps. First, we analyzed predicted $T_g$ 
distributions across a dataset of 48208 systematically substituted vinyl polymers spanning four families. Second, we performed pairwise statistical analyses over 12052 matched polymer pairs to quantify the magnitude and consistency of predicted $T_g$ shifts. Third, we compared these trends against experimentally measured $T_g$ values from three independently synthesized polymers and eight polymers drawn from the experimental literature, all absent from the fine-tuning dataset. Throughout this analysis, predictions were generated using the standalone Periodic-TDL model, without descriptor-based stacking (+Chem) or residual correction (+RC).

\subsubsection*{Pendant Group Chemistry Drives Systematic \texorpdfstring{$T_g$}{Tg} Shifts Across Polymer Families}
\begin{figure}[htp!]
    \centering
    \includegraphics[width=\linewidth]{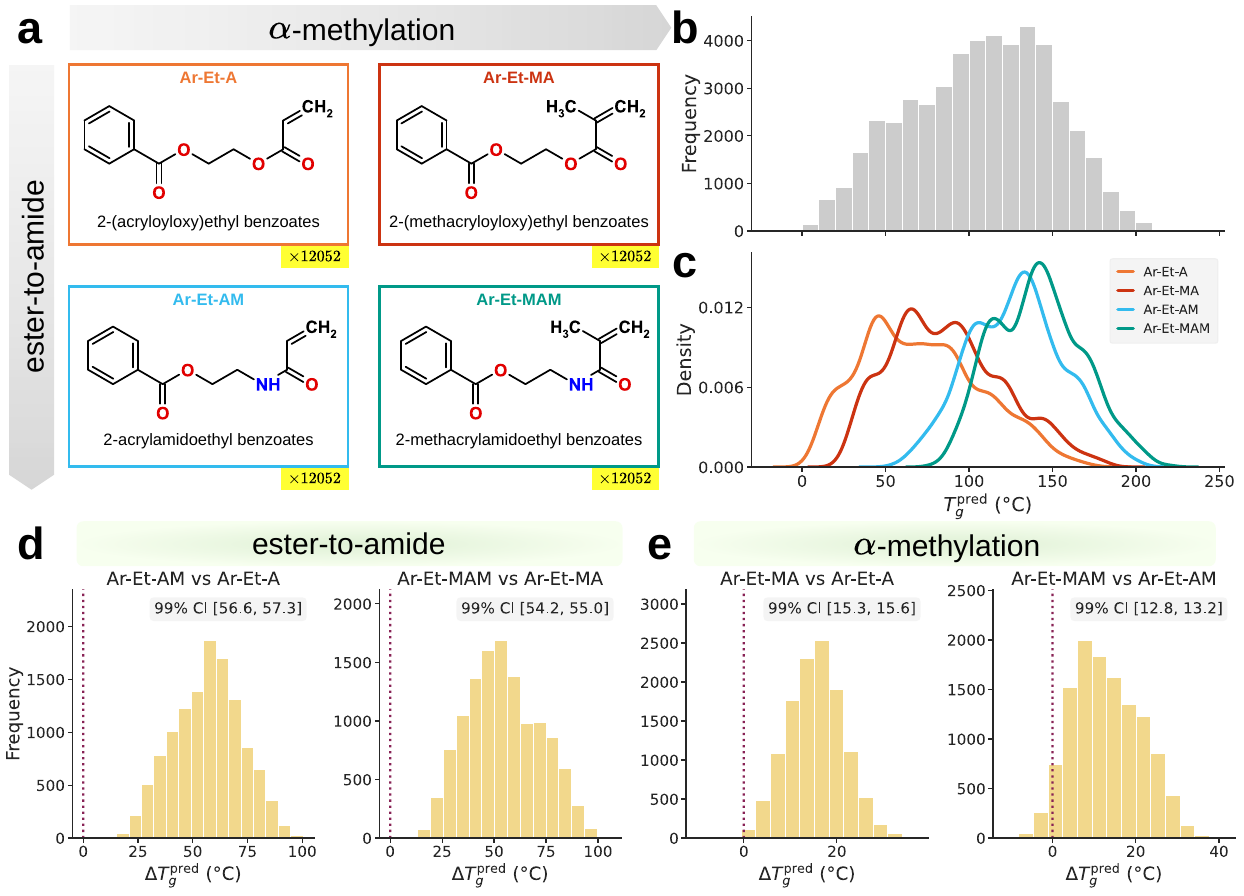}
    \caption{{\textbf{Predicted glass transition temperature trends for systematically substituted acrylate- and acrylamide-based polymers.}
    \textbf{(a)} Monomer structures for the four polymer families: poly(acrylate) (Ar-Et-A), poly(methacrylate) (Ar-Et-MA), poly(acrylamide) (Ar-Et-AM), and poly(methacrylamide) (Ar-Et-MAM), arranged to reflect the two structural modifications under study: ester-to-amide (vertical axis) and backbone $\alpha$-methylation (horizontal axis). Systematic substitution on the phenyl ring yields 12,052 monomers per family.
    \textbf{(b)} Overall distribution of predicted $T_g$ across the full dataset.
    \textbf{(c)} Stratified $T_g$ distributions for each polymer family, showing systematic shifts associated with ester-to-amide substitution and $\alpha$-methylation.
    \textbf{(d)} Distributions of pairwise $\Delta T_g^{\mathrm{pred}}$ for matched polymer pairs differing by ester-to-amide substitution, with $99\%$ confidence intervals.
    \textbf{(e)} Distributions of pairwise $\Delta T_g^{\mathrm{pred}}$ for matched polymer pairs differing by $\alpha$-methylation, with $99\%$ confidence intervals.}}    
    \label{fig:QualTrend}
    \end{figure}

We generated a computational dataset of 48208 monomer structures via systematic substitution on the phenyl ring of four monomers illustrated in Fig.~\ref{fig:QualTrend}(a). The dataset spans two ester-pendant families composed of substituted 2-(acryloyloxy)ethyl benzoates (Ar-Et-A) and 2-(methacryloyloxy)ethyl benzoates (Ar-Et-MA), and two amide-pendant families composed of substituted 2-acrylamidoethyl benzoates (Ar-Et-AM) and 2-methacrylamidoethyl benzoates (Ar-Et-MAM). Details of dataset generation are provided in Methods.

The overall distribution of predicted $T_g$ across the full dataset spans approximately $0\,^\circ\mathrm{C}$ to $225\,^\circ\mathrm{C}$, as shown in Fig.~\ref{fig:QualTrend}(b). When stratified by polymer family, distinct $T_g$ distributions emerge for each of the four families [Fig.~\ref{fig:QualTrend}(c)], with statistically significant differences confirmed across all comparisons ($p < 0.001$, see Methods). Mean $T_g$ values for each family are reported in \textbf{Table S6}.

We further examined polymer candidates within low ($T_g^{\mathrm{pred}} \leq 30\,^\circ\mathrm{C}$), medium ($100\,^\circ\mathrm{C} \leq T_g^{\mathrm{pred}} \leq 130\,^\circ\mathrm{C}$), and high ($T_g^{\mathrm{pred}} \geq 200\,^\circ\mathrm{C}$) predicted $T_g$ ranges, yielding 1702, 12029, and 221 polymers, respectively. Schuffenhauer Synthetic Accessibility (SA) scores \cite{ertl_estimation_2009} computed for the corresponding monomers indicate that higher $T_g$ ranges are associated with more synthetically challenging monomers (\textbf{Fig.~S4}). This is consistent with bulkier, more polar, and hydrogen-bonding-capable pendant groups contributing to elevated $T_g$, as commonly observed in polymers.

\subsubsection*{Amide Substitution and \texorpdfstring{$\alpha$}{alpha}-Methylation Produce Consistent \texorpdfstring{$T_g$}{Tg} Elevations}
We evaluated the effect of ester-to-amide substitution and backbone $\alpha$-methyl substitution on $T_g^{pred}$ via pairwise analyses over 12052 matched monomer pairs, isolating each structural modification while controlling for all other aspects of monomer structure. { Distributions of $\Delta T_g^{\mathrm{pred}}$ for ester-to-amide and $\alpha$-methylation comparisons are shown in Fig.~\ref{fig:QualTrend}(d) and Fig.~\ref{fig:QualTrend}(e) respectively, with $99\%$ confidence intervals computed from the empirical distribution. Statistical analysis details are provided in Methods.}

Replacing ester with amide pendant groups produced the largest and most consistent $T_g$ elevation. The predicted shift was positive across all matched pairs for both the acrylate--acrylamide comparison (mean $= 57.0\,^\circ\mathrm{C}$, $99\%$ CI: [56.6, 57.3]) and the methacrylate--methacrylamide comparison (mean $= 54.6\,^\circ\mathrm{C}$, $99\%$ CI: [54.2, 55.0]). Backbone $\alpha$-methyl substitution produced a smaller but statistically significant elevation in both the acrylate--methacrylate (mean $= 15.4\,^\circ\mathrm{C}$, $99\%$ CI: [15.3, 15.6]) and the acrylamide--methacrylamide comparisons (mean $= 13.0\,^\circ\mathrm{C}$, $99\%$ CI: [12.8, 13.2]), with a minority of matched pairs exhibiting negative $\Delta T_g^{\mathrm{pred}}$ in both cases. All four comparisons were statistically significant ($p < 0.001$, two-sided $t$-test, corrected for multiple comparisons).

\subsubsection*{Predicted Directional Trends Are Confirmed by Experimental Measurements}
To corroborate the trends identified above, we synthesized three polymers from the systematically created dataset, namely poly(2-(acryloyloxy)ethyl benzoate) (PBz-Et-A), poly(2-(methacryloyloxy)ethyl benzoate) (PBz-Et-MA), and poly(2-acrylamidoethyl benzoate) (PBz-Et-AM). All three polymers were absent from the fine-tuning dataset and have not been previously characterized in the experimental literature. Synthesis and characterization were carried out using identical experimental protocols, detailed in Supplementary Material. Table~\ref{tab:experimental_tg_validation} reports predicted and experimentally measured $T_g$ values for the three synthesized polymers. Two of the three predictions agree with experiment to within $4\,^\circ \mathrm{C}$ (PBz-Et-A and PBz-Et-MA), demonstrating that Periodic-TDL can achieve quantitative accuracy for novel polymers outside the fine-tuning dataset.

For polymer design, directional accuracy across structurally related polymers is a crucial and practically relevant criterion, as it evaluates whether the model captures meaningful structure--property trends rather than only minimizing absolute errors in the training set. The three synthesized polymers yield two matched 
pairs, PBz-Et-A versus PBz-Et-AM (ester-to-amide effect) and PBz-Et-A versus PBz-Et-MA ($\alpha$-methylation effect). To obtain a broader assessment spanning diverse pendant group chemistries, we supplement these with four additional polymer pairs drawn from the experimental literature, all absent from the fine-tuning dataset. Three pairs were selected preferentially from polymers bearing aromatic pendant groups, consistent with the chemical space of the systematically substituted dataset. One glycopolymer pair was also included to probe generalization to non-aromatic systems. Together, these yield six matched pairs covering both structural modifications under study. Table~\ref{tab:literature_delta_tg_validation} reports predicted and experimental $\Delta T_g$ values for all six pairs. Predicted directional trends were in agreement with experiment in all six cases, spanning both 
ester-to-amide substitution and $\alpha$-methylation effects across structurally diverse pendant group chemistries.

Overall, the analyses presented here establish the chemical credibility of Periodic-TDL predictions. Ester-to-amide substitution and backbone $\alpha$-methyl substitution each produce statistically significant $T_g$ elevations across matched polymer pairs. The predicted directional trends are also confirmed against experimental measurements across pairs spanning diverse pendant group chemistries.

\section*{Discussion}\label{sec:Discussion}
We introduce Periodic-TDL, a framework that encodes the periodic, many-body, and multiscale structure of polymers through a periodic Vietoris–Rips filtration. The Hierarchical Simplicial Message Passing encoder propagates information from higher-order simplices and long-range interactions to atom-level representations at the covalent scale. Following pretraining on one million unlabelled polymers and fine-tuning on labelled datasets, Periodic-TDL achieves competitive performance across electronic, optical, physical, and thermal property-prediction tasks. 
Among the polymer property prediction baselines considered in our work, PerioGT \cite{wu2025periodicity} is the most recent and is especially relevant because it also incorporates polymer periodicity into its modeling framework.
In PerioGT, however, an implicit periodic prior driven by chemical knowledge is incorporated through a contrastive pretraining step, whereas Periodic-TDL encodes periodicity explicitly in the construction of its periodic Vietoris–Rips complex.
Moreover, PerioGT augments its structural representation with global physicochemical descriptors and molecular fingerprints, whereas the core Periodic-TDL model is derived from atom, bond and simplex level structural features. Our results indicate that incorporating comparable global chemical information can further improve Periodic-TDL when sufficient training data is available. 
This finding suggests that physicochemical descriptors capture complementary chemical information that may not be fully recovered from local connectivity information alone.

Beyond predictive accuracy, we assessed the chemical credibility of Periodic-TDL by examining the changes in $T_g$ upon ester-to-amide substitution and backbone $\alpha$-methylation in acrylate- and acrylamide-based polymers. 
Both effects have been noted as phenomenological observations and documented experimentally for isolated polymer pairs, but their magnitude and consistency across a systematically varied family of polymers have not previously been quantified. 
Periodic-TDL predicts that (meth)acrylamides exhibit higher $T_g$ than (meth)acrylates, and methacrylates (methacrylamides) exhibit higher $T_g$ than the corresponding acrylates (acrylamides). 
Both trends were further validated against experimental measurements from six polymer pairs, including three newly synthesized polymers absent from literature. 
Next, we discuss the mechanistic origins of these trends and speculate on how specific features of the Periodic-TDL architecture may account for its sensitivity to individual functional group modifications.

The $T_g$ of polymers is influenced by the polarity, bulkiness, and flexibility of pendant groups. 
Specifically, the higher $T_g$ of (meth)acrylamides relative to (meth)acrylates has been attributed to two mechanisms. 
First, the partial double-bond character of the C--N bond increases pendant-group stiffness in (meth)acrylamides, whereas the C--O bond in (meth)acrylates allows free rotation \cite{gallardo_effect_1993}. 
Second, the acrylamide group can act as both a hydrogen-bond donor and acceptor, enabling physical cross-linking between amide and carbonyl groups \cite{yuan_multifunctional_2026, aydogan_ayaz_influence_2013}. 
The higher $T_g$ of methacrylates (methacrylamides) relative to acrylates (acrylamides) has been attributed to steric hindrance from the $\alpha$-methyl group, which restricts chain mobility \cite{zhou_high_2016, fleischhaker_glass-transition-_2014, adharis_synthesis_2018}. 
The agreement with known trends likely arises from two features of Periodic-TDL. 
First, the periodic Vietoris–Rips filtration at larger cutoffs, $\epsilon_2 = 3.0\,\text{\AA}$ and $\epsilon_3 = 4.0\,\text{\AA}$, incorporates interactions at length scales relevant to non-covalent contacts such as hydrogen bonding. 
Second, the additional $\alpha$-methyl group increases local steric density, leading to more higher-dimensional simplices at intermediate cutoffs. 
This structural difference is likely reflected in curvature-based simplex features, where more positive Forman--Ricci curvature indicates locally dense regions in the complex.

A distinctive feature of Periodic-TDL is its integration of multiscale interactions across filtration levels. Existing deep learning frameworks that incorporate filtration typically treat each cutoff independently and obtain a final molecular representation by pooling and concatenating features across cutoffs \cite{shen2023molecular, shen2026molecular, nguyen2026topology}. 
In contrast, the HSMP encoder processes all cutoffs within a unified hierarchical pipeline. 
Information propagates from coarser to finer spatial scales through cross-scale refinement, such that atom and bond representations at the covalent scale already incorporate contributions from higher-order simplices and long-range interactions. 
This design preserves compatibility with established graph-based pretraining workflows, allowing HSMP to replace conventional graph encoders without requiring new pretraining objectives. 
To our knowledge, this work is also the first to incorporate discrete curvature into a TDL framework. 
In particular, Forman--Ricci curvature provides a principled initial signal for simplices of any dimension, which is necessary for message passing to operate meaningfully on higher-order topological domains.

Several limitations of the current work point to promising future directions. 
First, Periodic-TDL is currently restricted to linear homopolymers and does not yet handle copolymers, for which additional information on stoichiometry and chain architecture would be required \cite{lin2019bigsmiles, aldeghi_graph_2022}. 
Second, the periodic Vietoris–Rips complex is constructed from cyclic permutations of a single repeating unit and therefore does not explicitly encode interactions between atoms in distant repeat units. 
Quotient complex constructions have been successfully applied to crystalline materials with three-dimensional periodicity \cite{hu2025quotient, you2026quotient}. 
However, many synthetic polymers are semicrystalline or amorphous rather than fully crystalline, and a careful assessment is therefore needed to determine whether such constructions can be transferred to polymer systems. 
Third, crosslinking is not represented in the current framework, despite being a structural factor that strongly influences $T_g$ \cite{nielsen1969cross, nakagawa2022star}. 
Finally, our computational analysis suggests that higher-$T_g$ polymers tend to be bulkier and more polar. This trend could be tested systematically through targeted synthesis of candidates selected across the predicted $T_g$ range.

\section*{Materials and Methods}\label{sec:Methods}
\subsection*{Construction of the Periodic Vietoris–Rips Complex}
We considered linear homopolymers composed of a single repeating unit arranged along an unbranched backbone. To construct a representation that was invariant to the arbitrary choice of the repeating unit, we generated all possible units following the fragmentation and rearrangement scheme proposed in Lo et al. \cite{lo_augmenting_2023}. Specifically, we first identified a backbone via the shortest path between the two terminal dummy atoms. Bonds along this backbone that were not part of rings were treated as breakable, yielding fragments that can be cyclically permuted while preserving chemical connectivity. Side-chain atoms remained attached to their corresponding backbone fragment. Each cyclic permutation produced a valid repeating unit with identical composition but shifted endpoints along the backbone. Connectivity was preserved across all permutations. 

Let $N$ be the number of atoms in the smallest unit and let the atom index set be $S = \{1,2,\dots,N\}$. Let $K$ denote the number of unique units generated using cyclic permutations. For each unique unit with index $k \in \{1,\dots,K\}$, let $e_\alpha^{(k)} \in \mathbb{R}^3$ denote the three-dimensional coordinate of atom $\alpha$. The periodic distance matrix $\mathbf{D} \in \mathbb{R}_{\ge 0}^{N \times N}$ was defined as
$$
\mathbf{D}_{\alpha \beta} = \min_{k} \, \| e_\alpha^{(k)} - e_\beta^{(k)} \|.
$$
The above construction aggregates geometric information across all cyclic permutations and yields a distance matrix that is invariant to any particular choice of repeating unit. Given the periodic distance matrix $\mathbf{D}$, the Vietoris--Rips complex at scale $\epsilon_i>0$ is defined as \cite{carlsson_topology_2009, edelsbrunner_computational_2010, otter_roadmap_2017}
$$
\mathrm{VR}_{\epsilon_i}(\mathbf{D})
=
\left\{
\sigma \subseteq S
\;\middle|\;
\mathbf{D}_{\alpha \beta} \le \epsilon_i \;\; \forall \alpha,\beta \in \sigma
\right\}.
$$
Thus, vertices correspond to atoms, edges correspond to atom pairs within distance $\epsilon$, and higher-dimensional simplices encode multi-body interactions among atoms whose pairwise distances do not exceed $\epsilon_i$. 

For the Periodic-TDL framework, we chose a nested filtration of Vietoris--Rips complexes at increasing spatial scales
$$
\mathrm{VR}_{\epsilon_1}(\mathbf{D})
\subset
\mathrm{VR}_{\epsilon_2}(\mathbf{D})
\subset
\mathrm{VR}_{\epsilon_3}(\mathbf{D}).
$$
Here, $\epsilon_1 = 2.0 \ \text{\AA}, \epsilon_2 = 3.0 \ \text{\AA}, \text{ and } \epsilon_3 = 4.0 \ \text{\AA}$. These cutoff values were fixed for all polymers used to train our model. The complex at $\epsilon_1=2$\,\AA\ operates at the scale of covalent bonds, while $\epsilon_2$ and $\epsilon_3$ progressively incorporate non-covalent interactions \cite{cang_representability_2018}. The pseudocode of the procedure used to construct the periodic distance matrix, and formal backgrounds on simplicial complexes and Vietoris--Rips filtration are provided in \textbf{Supplementary Material}.

\subsection*{Initialization of Simplicial Features Across Filtration Levels}
For each filtration level $\epsilon_i$, we assigned vertex (0-simplex), edge (1-simplex), and triangle (2-simplex) features that combine chemical descriptors with curvature-based geometric information. Chemical descriptors were invariant across filtration levels since they only capture properties associated with covalent bonds. On the other hand, curvature features were computed independently for each $\epsilon_i$, reflecting the scale-dependent topology induced by the cutoff parameter.
For a vertex $u$, the initial feature vector was defined as
$$
\mathbf{x}_{u,\epsilon_i}
=
\bigl[
\mathbf{x}^{\text{atom}}_u
\;\Vert\;
\mathbf{x}^{\text{curv}}_{u,\epsilon_i}
\bigr].
$$
For an edge $e$, 
$$
\mathbf{x}_{e,\epsilon_i}
=
\bigl[
\mathbf{x}^{\text{bond}}_e
\;\Vert\;
\mathbf{x}^{\text{curv}}_{e,\epsilon_i}
\bigr].
$$
For a 2-simplex $f$, only geometric information was assigned via curvature features,
$$
\mathbf{x}_{f,\epsilon_i}
=
\mathbf{x}^{\text{curv}}_{f,\epsilon_i}.
$$
All simplex features were subsequently mapped to a common hidden dimension via learnable linear projections prior to multi-head simplicial message passing.
\subsubsection*{Chemical Features}
Chemical descriptors were computed using RDKit \cite{greg_landrum_2023_7671152}. 
These encode local atomic and covalent bond properties and depend only on the underlying molecular graph derived from the repeating unit. 
Each vertex $u$ was assigned an atom-level feature vector $\mathbf{x}^{\mathrm{atom}}_u \in \mathbb{R}^{70}$ constructed from one-hot encodings of atomic properties, including element type, atomic degree, implicit valence, formal charge, number of radical electrons, hybridization state, and aromaticity. 
For edges corresponding to covalent bonds, a bond-level descriptor $\mathbf{x}^{\mathrm{bond}}_e \in \mathbb{R}^{6}$ was assigned, consisting of one-hot encodings of bond type along with binary indicators for conjugation and ring membership. 
If an edge in the Vietoris–Rips complex did not correspond to a covalent bond, its bond feature vector was set to the zero vector $\mathbf{x}^{\text{bond}}_e = \mathbf{0}$ in $\mathbb{R}^{6}$. This served as an explicit masking mechanism indicating the absence of covalent bond attributes, while geometric information was retained through curvature-based features.
\subsubsection*{Curvature Features}
Geometric features were derived from Forman’s discretization of curvature \cite{forman_bochners_2003, sreejith_forman_2016, sreejith_systematic_2017, samal_comparative_2018, wee2021forman}, which assigns a curvature function on simplices of any dimension. 
Given a Vietoris--Rips complex $\mathrm{VR}_{\epsilon_i}(\mathbf{D})$, curvature values were computed for vertices, edges, and triangles. 
We denote the curvature of a simplex $\sigma$ at threshold $\epsilon_i$ by $\mathcal{F}_{\epsilon_i}(\sigma)$.

To incorporate geometric information from intermediate interaction scales without explicitly performing message passing at all thresholds, we calculated Forman curvature over a finer set of Vietoris–Rips complexes around each base cutoff $\epsilon_i$. Specifically, for each $\epsilon_i$, additional Vietoris--Rips complexes were generated at thresholds
$$
\epsilon_i + k\Delta, 
\quad k = 0,1,2,3,4,
\qquad \Delta = 0.25\,\text{\AA}.
$$
Forman curvature was then computed independently on each resulting complex. For example, at the covalent-scale cutoff $\epsilon_1 = 2.0\,\text{\AA}$, curvature was evaluated at $\{2.0, 2.25, 2.5, 2.75, 3.0\}\,\text{\AA}$, yielding five curvature values per simplex.

The curvature-based feature vector for a simplex $\sigma$ at filtration level $\epsilon_i$ was defined as
$$
\mathbf{x}^{\mathrm{curv}}_{\sigma,\epsilon_i}
=
\bigl[
\mathcal{F}_{\epsilon_i}(\sigma),
\mathcal{F}_{\epsilon_i+\Delta}(\sigma),
\mathcal{F}_{\epsilon_i+2\Delta}(\sigma),
\mathcal{F}_{\epsilon_i+3\Delta}(\sigma),
\mathcal{F}_{\epsilon_i+4\Delta}(\sigma)
\bigr]
\in \mathbb{R}^{5}.
$$

Since Forman curvature may vary substantially in magnitude across complexes of different sizes, curvature values were normalized using a temperature-scaled sigmoid transformation
$$
f_T(x)
=
\frac{1}{1 + \exp(-x / T)},
\qquad T = 10,
$$
followed by centering to the interval $(-1,1)$ via $2f_T(x) - 1$. 
This transformation preserves the sign of curvature which encodes meaningful geometric information. For example, negative curvature often indicates bottleneck regions, whereas positive curvature corresponds to locally dense regions \cite{samal_comparative_2018, wee2021forman}. Moreover, it prevents excessively large curvature values from slowing convergence and compromising the numerical stability of gradient updates. 
Details on the computation of Forman curvature on simplicial complexes are provided in \textbf{Supplementary Material}.

\subsection*{Unlabelled and Labelled Polymer Datasets}
The HSMP architecture in Periodic-TDL serves as a geometric encoder that maps a polymer to its latent space representation. 
To learn representations that are transferable across polymer property prediction tasks, we adopt a two-stage training protocol consisting of large-scale pretraining on an unlabelled dataset followed by supervised fine-tuning on labelled downstream datasets.

For pretraining, we used the PI1M dataset \cite{ma_pi1m_2020}, a benchmark resource in polymer informatics comprising approximately one million polymers represented as pSMILES strings \cite{kuenneth_polybert_2023}. PI1M was constructed by training a generative model on polymers curated from the PolyInfo database and subsequently sampling $\sim$1M synthetic polymers, thereby providing broad chemical diversity for representation learning. Each entry in PI1M corresponds to a monomer repeating unit encoded in pSMILES format, where the symbol `*' denotes dummy atoms indicating polymerization sites.
We finetuned the HSMP encoder on nine regression datasets corresponding to polymer property prediction tasks: 
$E_{gc}$ (bandgap, chain) \cite{kamal_novel_2021}, 
$E_{ib}$ (electron injection barrier) \cite{kamal_novel_2021}, 
$E_{gb}$ (bandgap, bulk) \cite{kuenneth_polymer_2021}, 
$E_{ea}$ (electron affinity) \cite{kuenneth_polymer_2021}, 
$E_i$ (ionization energy) \cite{kuenneth_polymer_2021}, 
EPS (dielectric constant) \cite{kuenneth_polymer_2021}, 
$N_c$ (refractive index) \cite{kuenneth_polymer_2021},
$X_c$ (crystallization tendency) \cite{kuenneth_polymer_2021}, and
$T_g$ (glass transition temperature) \cite{malashin_estimation_2024}. 
These targets span thermal, electronic, optical, and physical property classes. 
Electronic ($E_{gc}, E_{ib}, E_{gb}, E_{ea}, E_i$), optical (EPS, $N_c$), and physical ($X_c$) properties were derived from density functional theory calculations, whereas $T_g$ corresponds to experimentally measured values. 
For each task, the original dataset consists of pSMILES strings of monomer repeating units paired with scalar property values. 
Summary statistics, including the number of samples, mean, standard deviation, and empirical range for each dataset, are reported in \textbf{Table S1}.

Both the unlabelled and labelled datasets provide polymer structures in pSMILES format, whereas the HSMP encoder operates on a periodic Vietoris–Rips filtration derived from three-dimensional atomic coordinates.  
For each polymer, we therefore generated 3D coordinates using RDKit by embedding the monomer and performing geometry optimization with the Universal Force Field (UFF) \cite{rappe_uff_1992}. 
For every pSMILES string, we generated all equivalent monomer representations using the cyclic rearrangement scheme described earlier (also see \textbf{Supplementary Material}), and constructed their corresponding UFF-optimized 3D atomic coordinates prior to computing the periodic distance matrix.
Since Periodic-TDL is designed for linear homopolymers, we retained only polymers with exactly two dummy atoms (`*') in their pSMILES representation. 
Polymers for which UFF optimization failed were excluded. 
The resulting polymers in each dataset were used for pretraining or for five-fold cross-validation during supervised fine-tuning.

\subsection*{HSMP Encoder Configuration}
The HSMP encoder within the Periodic-TDL framework contains multi-head simplicial message passing modules. These modules were fixed with a hidden dimension of $768$ and $12$ message-passing heads. Message passing was performed hierarchically across filtration levels with 4 edge-update layers followed by 6 node-update layers at the coarsest level ($\epsilon_3$), 4 edge-update layers followed by 6 node-update layers at the intermediate level ($\epsilon_2$), and 6 node-update layers at the finest level ($\epsilon_1$). At the finest filtration level, the encoder produced atom-level and bond-level representations. 
Further, a polymer-level representation was obtained via global mean pooling of atom representations.

\subsection*{Pretraining Details}
Our pretraining task consists of three self-supervised tasks defined on the atom, bond, and polymer level representations respectively: atom context prediction, bond context prediction, and functional-group (FG) prediction. These objectives were adapted from the molecular pretraining approach introduced in GROVER \cite{rong_self-supervised_2020} to operate on periodic Vietoris–Rips filtration of polymers. 
For atom and bond context prediction, each atom and bond was assigned a discrete context string encoding its 1-hop local chemical environment under the canonical scheme described in \cite{rong_self-supervised_2020}. 
The context strings were treated as categorical labels. 
The resulting vocabularies contained 4,580 atom context classes and 6,775 bond context classes. 
For each polymer, a random subset of atoms and bonds were selected as prediction targets with sampling fractions $p_{\text{atom}} = 0.15$ and $p_{\text{bond}} = 0.15$. 
Polymer-level supervision was incorporated through FG prediction, where a fixed set of 85 functional groups were defined using RDKit fragment descriptors. 
Since multiple functional groups may be present within a single polymer, FG prediction was formulated as a multi-label classification task.

The overall pretraining objective was defined as a weighted sum of the three losses:
$$
\mathcal{L}_{\text{total}}
=
w_{\text{atom}} \mathcal{L}_{\text{atom}}
+
w_{\text{bond}} \mathcal{L}_{\text{bond}}
+
w_{\text{fg}} \mathcal{L}_{\text{fg}},
$$
where $\mathcal{L}_{\text{atom}}$ and $\mathcal{L}_{\text{bond}}$ are categorical cross-entropy losses, and $\mathcal{L}_{\text{fg}}$ is binary cross-entropy. 
The loss weights were fixed to $w_{\text{atom}} = 2.0$, $w_{\text{bond}} = 1.0$, and $w_{\text{fg}} = 5.0$ in all experiments. 
During pretraining, 95\% of the PI1M dataset was used for optimization and 5\% was held out for validation. 
Models were optimized using AdamW \cite{loshchilov_decoupled_2019} with learning rates of $2\times10^{-4}$ for the encoder and $10^{-3}$ for the task-specific prediction heads. 
A weight decay of $10^{-4}$ was applied to all parameters, and gradients were clipped using a threshold of 5.0 \cite{pascanu_difficulty_2013}. 
No dropout was applied during pretraining. 
Training was performed for 10 epochs with a batch size of 64. 
Model checkpoints were saved after each epoch, and the final model was selected based on the lowest validation loss. 
The selected checkpoint achieved a training loss of 0.0763 and a validation loss of 0.0593.

\subsection*{Finetuning Details}
As the final stage of the Periodic-TDL framework, the pretrained HSMP encoder was fine-tuned on nine supervised polymer property prediction tasks. 
The polymer-level representation was obtained by global mean pooling of the learned atom embeddings. 
A two-layer regression head was then applied to this representation. The regression head consisted of a linear layer mapping the 768-dimensional polymer representation to a 768-dimensional hidden representation, followed by layer normalization, a $ReLU$ activation, dataset-specific dropout, and a final linear layer producing a scalar prediction.

All models were optimized using mean squared error (MSE) loss. Model performance was evaluated using root mean squared error (RMSE) and the coefficient of determination ($R^2$). 
Validation RMSE was used exclusively for checkpoint selection, with the final model corresponding to the checkpoint that achieved the lowest validation RMSE across the complete training trajectory.

Fine-tuning was performed using a two-stage optimization procedure. 
During the first stage, the pretrained encoder was frozen and only the regression head was optimized for 10 epochs \cite{yosinski_how_2014}. 
The regression head was trained using AdamW \cite{loshchilov_decoupled_2019} with a learning rate of $3\times10^{-4}$ and the dataset-specific weight decay described below. 
A cosine annealing schedule was applied during this stage, with the learning rate reduced to a minimum of $3\times10^{-5}$ over the 10 warm-up epochs.

During the second stage, the encoder was unfrozen and the encoder and regression head were jointly optimized. 
The encoder learning rate was set to $1\times10^{-4}$ for all datasets except $X_c$, for which it was set to $2\times10^{-4}$. 
The regression-head learning rate was fixed at $1\times10^{-3}$ for all datasets. 
A cosine annealing schedule with warm restarts was used during joint optimization, with restart cycles of 10 epochs, a cycle multiplier of 1, and a minimum learning rate of $1\times10^{-6}$ \cite{loshchilov_sgdr_2017}. 
Gradients were clipped to a maximum global norm of 5.0 \cite{pascanu_difficulty_2013}.
The second stage was run for 40 epochs for $T_g$, 50 epochs for $E_{gc}$, and 60 epochs for $E_{ib}$, $E_{gb}$, $E_{ea}$, $E_i$, EPS, $N_c$, and $X_c$. 

Validation RMSE was evaluated after every epoch in both stages, and the lowest-validation-RMSE checkpoint across the two stages was retained. The test set was evaluated only after restoring this selected checkpoint. 
Target standardization was applied for $T_g$, $E_{gc}$, $E_{ib}$, $E_{gb}$, $E_{ea}$, $E_i$, EPS, and $N_c$. All experiments used predefined five-fold cross-validation splits shared between Periodic-TDL and the baseline models. For each fold, 80\% of the data was assigned to the outer training partition and 20\% was held out as the test partition. 20\% of the training samples were reserved for validation. 
A batch size of 24 was used for all datasets. Weight decay was set to 0 for $T_g$ and $E_{gc}$, 0.05 for $E_{ib}$, 0.02 for $E_{gb}$, $E_{ea}$, $E_i$, EPS, and $N_c$, and $1\times10^{-4}$ for $X_c$. 
Regression-head dropout was set to 0 for $T_g$, $E_{gc}$, and $E_{ib}$; 0.1 for $E_{gb}$, $E_{ea}$, $E_i$, EPS, and $N_c$; and 0.3 for $X_c$. Encoder embedding dropout was set to 0 for all datasets.

\subsection*{Inclusion of Additional Chemical Information and Residual Correction}
To incorporate complementary chemical information, we developed Periodic-TDL (+Chem). Separate predictors were trained using the frozen pretrained Periodic-TDL representation, two-dimensional Mordred descriptors, and Morgan fingerprints. Mordred descriptors were modeled using both a multilayer perceptron and an XGBoost regressor, whereas the frozen Periodic-TDL representations and Morgan fingerprints were modeled using multilayer perceptrons. Their predictions were combined with those of the fully fine-tuned Periodic-TDL model using a regularized linear stacking model with Huber loss, which reduces sensitivity to samples with unusually large prediction errors. The descriptor-based predictors were trained on the training split, and the stacking model was fitted using the validation split of each fold.

Periodic-TDL (+Chem, +RC) further applies a similarity-based residual correction to the stacked prediction. Cross-fitted validation predictions were used to estimate the residual errors of the stacking model. For each test polymer, a correction was derived from validation polymers with similar Periodic-TDL pretrained embeddings and Morgan fingerprints, and the estimates from the two similarity spaces were averaged and added to the stacked prediction. This approach was inspired by ResMem \cite{yang2023resmem}, which showed that a parametric model can capture the broadly generalizable component of a target function, while a similarity-based memory of its residual errors can recover complementary information and reduce test error further \cite{yang2023resmem}. All residual estimates were constructed exclusively from the validation data within each fold.

\subsection*{Systematically Substituted Polymer Dataset}
We generated a comprehensive dataset of systematically substituted polymers to evaluate structure-property relationships in glass transition temperature. Four unsubstituted monomer scaffolds were selected as base structures: 2-(acryloyloxy)ethyl benzoate (Ar--Et--A), 2-(methacryloyloxy)ethyl benzoate (Ar--Et--MA), 2-acrylamidoethyl benzoate (Ar--Et--AM), and 2-methacrylamidoethyl benzoate (Ar--Et--MAM). These scaffolds represent the four backbone families poly(acrylate), poly(methacrylate), poly(acrylamide), and poly(methacrylamide), enabling direct comparison of ester versus amide functional groups and the presence versus absence of $\alpha$-methyl substitution.

For each scaffold, systematic substitution was performed on the phenyl ring at the ortho, meta, and para positions. We considered four substitution patterns: (i) no substituent, (ii) mono-substitution, (iii) di-substitution, and (iv) tri-substitution. At each substituted position, we assigned one of thirteen functional groups including alkyl and halogens. All substituted monomer structures were generated and subsequently converted to pSMILES strings. This procedure yielded 12,052 unique monomers per family and 48,208 polymers total across the four families. 

\subsubsection*{Glass Transition Temperature Prediction}
To obtain predicted $T_g$ values for the systematically substituted dataset, we applied considered the experimental $T_g$ dataset used for benchmarking Periodic-TDL. Each polymer was converted to its periodic Vietoris–Rips complex representation and processed through the HSMP encoder. The training data was randomly divided into ten folds. For each fold, the HSMP encoder was trained on nine folds while the tenth fold was used for early stopping. This yielded ten models optimized on disjoint validation splits. Importantly, none of the generated or validation polymers were used during training.

For inference on the generated dataset, literature polymers, and newly synthesized polymers, each polymer pSMILES was processed through the HSMP encoder using all ten trained models. The predicted $T_g$ for each polymer was obtained by averaging the outputs from the ten models, with uncertainty quantified as the standard deviation across the ensemble. Predicted values are reported as mean $\pm$ standard deviation.

\subsubsection*{Statistical Analysis of Predicted Glass Transition Temperatures}
Family-wise $T_g$ distributions were characterized by computing mean and standard deviation for each of the four polymer families (Ar--Et--A, Ar--Et--MA, Ar--Et--AM, Ar--Et--MAM). To test for significant differences between families, we applied two-sample Mann--Whitney U tests to each pair of families using a two-sided alternative hypothesis. The resulting p-values were corrected using the Holm method to control the family-wise error rate, and differences with corrected $p < 0.001$ were considered statistically significant.

To isolate the effects of specific structural modifications, we constructed matched polymer pairs across families using the recorded ring-substitution patterns and substituent identities. Four pairwise comparisons were performed: (i) Ar--Et--AM versus Ar--Et--A (ester-to-amide), (ii) Ar--Et--MAM versus Ar--Et--MA (ester-to-amide), (iii) Ar--Et--MA versus Ar--Et--A ($\alpha$-methylation), and (iv) Ar--Et--MAM versus Ar--Et--AM ($\alpha$-methylation). We tested whether mean $\Delta T_g$ deviated significantly from zero using two-sided one-sample $t$-tests. Confidence intervals (99\%) were calculated from the $t$-distribution. 

\subsection*{Literature Validation, Experimental Synthesis and Characterization}
To validate predicted trends against experimental observations, we identified published reports describing the effect of replacing ester with amide functional groups or adding $\alpha$-methyl substitution in acrylic polymers. We selected studies with matched structures consistent with our computational analyses and reliable experimental $T_g$ values. This yielded 8 polymers comprising 4 matched pairs. For each literature polymer, the monomer structure was converted to a pSMILES string and the predicted $T_g$ was obtained using the ensemble procedure described above. Polymers were grouped into pairs matching the literature structural modifications, and we computed $\Delta T_g^{(\text{pred})}$ and $\Delta T_g^{(\text{lit})}$ for each pair. Directional agreement was assessed by determining whether the predicted and experimental $\Delta T_g$ values shared the same sign.

To provide independent experimental validation, we synthesized three polymers that were entirely absent from the training dataset and had not been previously characterized in the experimental literature, namely poly(2-(acryloyloxy)ethyl benzoate), poly(2-(methacryloyloxy)ethyl benzoate), and poly(2-acrylamidoethyl benzoate). These polymers were selected to enable direct evaluation of both ester-to-amide and $\alpha$-methylation effects. Detailed synthesis and characterization conditions are provided in the Supplementary Material. Predicted $T_g$ values were compared with experimental measurements to assess model accuracy on independently synthesized materials.

To assess the feasibility of synthesizing the generated polymer library, we computed Schuffenhauer Synthetic Accessibility (SA) scores for all 48,208 monomers. SA scores range from 1 (highly synthesizable) to 10 (very difficult to synthesize). Polymers were categorized into three $T_g$ ranges: low ($T_g \leq 30\,^\circ$C), medium ($100\,^\circ$C $\leq T_g \leq 130\,^\circ$C), and high ($T_g \geq 200\,^\circ$C). For each range, we examined the distribution of SA scores and identified monomers with the lowest SA scores as promising synthetic targets.

\clearpage

\clearpage
\begin{table}[htp!]
\centering
\renewcommand{\arraystretch}{2.0}
\setlength{\tabcolsep}{6pt}
\scriptsize
\caption{\textbf{Test RMSE of Periodic-TDL and baseline models across nine polymer property prediction tasks.}
Values are reported as mean $\pm$ standard deviation across five test folds, with all models evaluated using identical data splits.
Periodic-TDL (+Chem) combines Periodic-TDL with global chemical descriptor-based predictors through linear stacking, and Periodic-TDL (+Chem, +RC) further incorporates a local residual correction based on polymer similarity.
Lower values indicate better performance.
The best and second-best mean values for each target are shown in \textbf{bold} and \underline{underlined}, respectively.}
\label{tab:rmse_comparison}
\resizebox{\textwidth}{!}{%
\begin{tabular}{lccccccccc}
\toprule
\textbf{Model} &
\makecell{$E_{ea}$ (eV)} &
\makecell{$E_i$ (eV)} &
\makecell{EPS} &
\makecell{$N_c$} &
\makecell{$X_c$ (\%)} &
\makecell{$E_{gb}$ (eV)} &
\makecell{$E_{ib}$ (eV)} &
\makecell{$E_{gc}$ (eV)} &
\makecell{$T_g$ ($^\circ$C)} \\
\midrule

Morgan (NN) &
0.431 $\pm$ 0.018 &
0.720 $\pm$ 0.047 &
0.737 $\pm$ 0.034 &
0.167 $\pm$ 0.008 &
19.41 $\pm$ 0.97 &
0.860 $\pm$ 0.042 &
0.642 $\pm$ 0.073 &
0.559 $\pm$ 0.010 &
37.99 $\pm$ 1.27 \\

polyBERT &
0.333 $\pm$ 0.009 &
0.443 $\pm$ 0.066 &
0.626 $\pm$ 0.048 &
0.113 $\pm$ 0.006 &
19.74 $\pm$ 0.82 &
0.626 $\pm$ 0.058 &
0.574 $\pm$ 0.081 &
0.483 $\pm$ 0.020 &
36.33 $\pm$ 1.89 \\

TransPolymer &
0.340 $\pm$ 0.042 &
0.454 $\pm$ 0.068 &
0.650 $\pm$ 0.032 &
0.117 $\pm$ 0.014 &
19.19 $\pm$ 0.70 &
0.609 $\pm$ 0.067 &
0.589 $\pm$ 0.080 &
0.462 $\pm$ 0.032 &
35.81 $\pm$ 2.17 \\

polyGNN &
0.309 $\pm$ 0.031 &
0.584 $\pm$ 0.144 &
0.631 $\pm$ 0.112 &
0.109 $\pm$ 0.011 &
21.23 $\pm$ 1.45 &
0.597 $\pm$ 0.041 &
0.577 $\pm$ 0.059 &
0.524 $\pm$ 0.040 &
43.35 $\pm$ 0.74 \\

MolCLR (GCN) &
0.391 $\pm$ 0.039 &
0.550 $\pm$ 0.074 &
0.629 $\pm$ 0.073 &
0.118 $\pm$ 0.011 &
20.66 $\pm$ 1.01 &
0.702 $\pm$ 0.098 &
0.573 $\pm$ 0.076 &
0.555 $\pm$ 0.018 &
44.10 $\pm$ 1.99 \\

MolCLR (GIN) &
0.341 $\pm$ 0.029 &
0.564 $\pm$ 0.126 &
0.641 $\pm$ 0.065 &
0.121 $\pm$ 0.010 &
20.77 $\pm$ 1.32 &
0.627 $\pm$ 0.078 &
0.549 $\pm$ 0.067 &
0.484 $\pm$ 0.022 &
41.53 $\pm$ 1.08 \\

TransChem &
0.376 $\pm$ 0.056 &
0.504 $\pm$ 0.153 &
0.663 $\pm$ 0.084 &
0.114 $\pm$ 0.007 &
19.64 $\pm$ 0.69 &
0.656 $\pm$ 0.061 &
0.637 $\pm$ 0.081 &
0.483 $\pm$ 0.031 &
35.72 $\pm$ 1.39 \\

MMPolymer &
0.342 $\pm$ 0.028 &
0.493 $\pm$ 0.089 &
0.691 $\pm$ 0.059 &
0.120 $\pm$ 0.015 &
19.33 $\pm$ 1.59 &
0.596 $\pm$ 0.056 &
0.606 $\pm$ 0.053 &
0.448 $\pm$ 0.019 &
35.77 $\pm$ 1.98 \\

PerioGT &
0.286 $\pm$ 0.026 &
0.440 $\pm$ 0.076 &
0.555 $\pm$ 0.064 &
0.098 $\pm$ 0.012 &
\second{17.92 $\pm$ 0.99} &
0.583 $\pm$ 0.094 &
0.545 $\pm$ 0.080 &
\second{0.433 $\pm$ 0.020} &
\second{32.13 $\pm$ 1.59} \\

\midrule

Periodic-TDL &
\best{0.276 $\pm$ 0.021} &
\best{0.391 $\pm$ 0.042} &
\best{0.511 $\pm$ 0.033} &
\best{0.089 $\pm$ 0.004} &
\best{17.84 $\pm$ 1.20} &
\second{0.545 $\pm$ 0.071} &
\second{0.543 $\pm$ 0.070} &
0.434 $\pm$ 0.022 &
34.14 $\pm$ 1.72 \\

Periodic-TDL (+Chem) &
0.283 $\pm$ 0.018 &
0.410 $\pm$ 0.046 &
0.516 $\pm$ 0.043 &
\second{0.092 $\pm$ 0.005} &
18.01 $\pm$ 0.93 &
0.547 $\pm$ 0.073 &
\best{0.527 $\pm$ 0.074} &
0.437 $\pm$ 0.023 &
32.73 $\pm$ 1.83 \\

Periodic-TDL (+Chem, +RC) &
\second{0.282 $\pm$ 0.018} &
\second{0.407 $\pm$ 0.050} &
\second{0.513 $\pm$ 0.065} &
\second{0.092 $\pm$ 0.009} &
18.33 $\pm$ 0.88 &
\best{0.540 $\pm$ 0.061} &
0.559 $\pm$ 0.067 &
\best{0.428 $\pm$ 0.018} &
\best{31.91 $\pm$ 1.64} \\

\bottomrule
\end{tabular}
}
\end{table}
\begin{table}[htp]
\centering
\renewcommand{\arraystretch}{1.3}
\setlength{\tabcolsep}{6pt}
\caption{\textbf{Experimental validation of predicted $T_g$ for newly synthesized polymers.} Three polymers absent from the fine-tuning dataset were synthesized and characterized using identical experimental protocols (see Supplementary Material). $T_g^{\mathrm{exp}}$ represents experimentally measured values. $T_g^{\mathrm{pred}}$ represents mean $\pm$ SD across ten cross-validation folds as predicted by Periodic-TDL.}
\label{tab:experimental_tg_validation}
\begin{tabular}{llcc}
\toprule
\textbf{Abbrev.} & \textbf{Polymer} & 
\textbf{$T_g^{exp}$ ($^\circ$C)} & 
\textbf{$T_g^{pred}$ ($^\circ$C)} \\
\midrule
PBz-Et-A  & poly(2-(acryloyloxy)ethyl benzoate)       
          & 7.3  & $7.4 \pm 4.5$  \\
PBz-Et-MA & poly(2-(methacryloyloxy)ethyl benzoate)   
          & 32.7 & $29.1 \pm 4.4$ \\
PBz-Et-AM & poly(2-acrylamidoethyl benzoate)          
          & 38.6 & $60.3 \pm 5.7$ \\
\bottomrule
\end{tabular}
\end{table}
\begin{table}[htp]
\centering
\renewcommand{\arraystretch}{2.0}
\setlength{\tabcolsep}{6pt}
\scriptsize
\caption{{\textbf{Pairwise validation of predicted $\Delta T_g$ trends.} Six polymer pairs, each differing by a single structural modification (ester-to-amide substitution or backbone $\alpha$-methylation), are evaluated for directional agreement between predicted and experimental $\Delta T_g$ values. Pairs marked as $^\dagger$ were synthesized in this work while remaining pairs were drawn from the experimental literature. All polymers were absent from the fine-tuning dataset. $\Delta T_g^{\mathrm{exp}}$ denotes experimentally measured values. $\Delta T_g^{\mathrm{pred}}$ denotes mean $\pm$ SD across ten cross-validation folds. Directional agreement indicates whether the sign of $\Delta T_g^{\mathrm{pred}}$ matches the sign of $\Delta T_g^{\mathrm{exp}}$. Full polymer names and individual $T_g$ values are provided in Supplementary \textbf{Table~S7}.}}
\label{tab:literature_delta_tg_validation}
\resizebox{\textwidth}{!}{%
\begin{tabular}{llccc}
\toprule
\textbf{Modification} & \textbf{Polymer pair} & 
\textbf{$\Delta T_g^{exp}$ ($^\circ$C)} & 
\textbf{$\Delta T_g^{pred}$ ($^\circ$C)} & 
\textbf{Directional agreement} \\
\midrule
Ester-to-amide  
  & PBz-Et-A $\rightarrow$ PBz-Et-AM $^\dagger$ 
  & 31.3 & $52.9 \pm 5.4$ & \cmark \\
Ester-to-amide  
  & Poly-O $\rightarrow$ Poly-A \cite{gallardo_effect_1993}          
  & 97.0 & $8.3 \pm 5.3$  & \cmark \\
\midrule
$\alpha$-methylation  
  & PBz-Et-A $\rightarrow$ PBz-Et-MA $^\dagger$                       
  & 25.4 & $21.7 \pm 3.3$ & \cmark \\
$\alpha$-methylation  
  & PSA $\rightarrow$ PSMA \cite{zhou_high_2016}                     
  & 10.0 & $11.1 \pm 1.9$ & \cmark \\
$\alpha$-methylation  
  & PVA $\rightarrow$ PVMA \cite{zhou_high_2016}                     
  & 15.0 & $17.3 \pm 3.1$ & \cmark \\
$\alpha$-methylation  
  & PGlc-$\beta$-EAAM $\rightarrow$ PGlc-$\beta$-EMAAM 
    \cite{adharis_synthesis_2018}                                     
  & 23.5 & $4.2 \pm 1.9$  & \cmark \\
\bottomrule
\end{tabular}}
\end{table}

\clearpage
\section*{Data and Code Availability}
The pretraining dataset (PI1M~\cite{ma_pi1m_2020}) and the downstream fine-tuning datasets~\cite{kamal_novel_2021, kuenneth_polymer_2021, malashin_estimation_2024} used in this work are made available in their respective original publications. 
The systematically substituted polymer dataset of 48208 structures along with predicted $T_g$ values across all ten cross-validation folds, and the experimental characterization data for the three independently synthesized polymers along with their corresponding predicted $T_g$ values, are publicly available at \url{https://github.com/yasharthy/Periodic-TDL}. 
All code for the construction of the periodic Vietoris–Rips complex, the Hierarchical Simplicial Message Passing (HSMP) encoder, and the pretraining and fine-tuning pipelines is publicly available at \url{https://github.com/yasharthy/Periodic-TDL}.

\begin{acknowledgments}
This work was supported in part by the Singapore Ministry of Education Academic Research fund Tier 1 grant RG16/23, Tier 2 grants MOE-T2EP20125-0004 and MOE-T2EP50223-0036. 
The computational work for this article was partially performed on resources of the National Supercomputing Centre (NSCC), Singapore (\url{https://www.nscc.sg}).
Y.Y. would like to thank Subhakanta Das, Longlong Li, Thanya Ramanathan, Cong Shen and Yipeng Zhang for discussions.
\end{acknowledgments}

\section*{Author contributions}
K.X. conceived and designed the research. Y.Y. performed the computational work and simulations. T.K.G.E. conducted the experiments. All authors contributed to writing and revising the manuscript.

\section*{Competing interests}
There are no competing interests to declare.

\bibliographystyle{unsrt}
\bibliography{refs}
\clearpage
\appendix
\setcounter{table}{0}
\renewcommand{\thetable}{S\arabic{table}}
\setcounter{figure}{0}
\renewcommand{\thefigure}{S\arabic{figure}}
\setcounter{equation}{0}
\renewcommand{\theequation}{S\arabic{equation}}
\makeatletter
\renewcommand{\theHtable}{SI.\arabic{table}}
\renewcommand{\theHfigure}{SI.\arabic{figure}}
\renewcommand{\theHequation}{SI.\arabic{equation}}
\makeatother

\begin{center}
{\Large\bfseries SUPPLEMENTARY INFORMATION\par}
\vspace{0.8em}
{\large\ for \par}
\vspace{0.8em}
{\large\bfseries
Periodic Topological Deep Learning for Polymer Design and Discovery
\par}
\vspace{1em}
Yasharth Yadav$^{1,*}$,
Tze Kwang Gerald Er$^{2}$,
Atsushi Goto$^{2}$,
and Kelin Xia$^{1,*}$ \\
\vspace{0.6em}
{\small
$^{1}$School of Physical and Mathematical Sciences, Nanyang Technological University, Singapore 637371
\par
$^{2}$School of Chemistry, Chemical Engineering and Biotechnology (CCEB), Nanyang Technological University, Singapore 637371
\par
}
\vspace{0.8em}
{\small
$^{*}$To whom correspondence should be addressed:
yasharth001@e.ntu.edu.sg; xiakelin@ntu.edu.sg
\par}
\end{center}
\clearpage

\section*{Computational Procedure for Periodic Distance Matrix Construction}\label{appx:A}
\noindent\rule{\linewidth}{0.8pt}
\noindent\textbf{Algorithm 1: Construction of the Periodic Distance Matrix} \\
\noindent\rule{\linewidth}{0.8pt}
{
    \renewcommand{\baselinestretch}{1.2}\selectfont
    \begin{algorithmic}[1]
    \Statex \textbf{Input:} Molecular structure of a single repeating unit
    
    \noindent\rule{\linewidth}{0.2pt}
    \Statex \textbf{Part I: Generation of translated repeating units}
    
    \State Identify backbone path between terminal dummy atoms
    \State Fragment along backbone bonds that are not part of rings
    \State Generate all cyclic permutations of backbone fragments
    \State Remove duplicate structures to obtain $K$ unique repeating units
    \Statex
    \For{$k = 1$ to $K$}
        \State Generate 3D coordinates $\{e_\alpha^{(k)}\}_{\alpha=1}^N$ using Universal Force Field
    \EndFor
    \Statex
    
    \noindent\rule{\linewidth}{0.2pt}
    \Statex \textbf{Part II: Periodic distance matrix computation}
    \Statex
    \For{$\alpha = 1$ to $N$}
        \For{$\beta = 1$ to $N$}
            \State $D_{\alpha \beta} \gets \min_{k} \| e_\alpha^{(k)} - e_\beta^{(k)} \|$
        \EndFor
    \EndFor
    
    \noindent\rule{\linewidth}{0.2pt}
    \Statex \textbf{Output:} Periodic distance matrix $\mathbf{D}$
    \end{algorithmic}
    \noindent\rule{\linewidth}{0.8pt}
    }

\section*{Mathematical Background on Simplicial Complexes and Vietoris--Rips Filtration}\label{appx:B}
\subsection*{Simplicial complexes}
Let $V$ be a finite set. A \emph{simplicial complex} $K$ on $V$ is a collection of non-empty subsets of $V$ satisfying:

\begin{enumerate}
    \item If $\sigma \in K$ and $\tau \subseteq \sigma$, then $\tau \in K$.
    \item For every $v \in V$, the singleton $\{v\}$ belongs to $K$.
\end{enumerate}

The elements of $V$ are called \emph{vertices}, and the elements of $K$ are called \emph{simplices}.  
A simplex $\sigma \in K$ containing $p+1$ vertices is called a \emph{$p$-simplex} and is said to have dimension $p$.  
If $\tau \subseteq \sigma$, then $\tau$ is called a \emph{face} of $\sigma$.  
The set of all $p$-simplices of $K$ is denoted by $K_p$.  
The dimension of $K$ is the maximum dimension among its simplices.

Simplicial complexes provide a combinatorial framework for encoding higher-order relationships. Vertices represent individual entities, edges represent pairwise relations, triangles encode three-body interactions, and higher-dimensional simplices capture multi-body interactions.

\subsection*{Vietoris--Rips complex}

Let $S=\{1,2,\dots,N\}$ be a finite set, and let 
\[
\mathbf{D} \in \mathbb{R}_{\ge 0}^{N \times N}
\]
be a symmetric distance matrix with $\mathbf{D}_{\alpha\alpha}=0$ for all $\alpha \in S$.  
For a threshold $\epsilon > 0$, the \emph{Vietoris--Rips complex} at scale $\epsilon$, denoted $\mathrm{VR}_\epsilon(\mathbf{D})$, is defined as

\[
\mathrm{VR}_\epsilon(\mathbf{D})
=
\left\{
\sigma \subseteq S
\;\middle|\;
\mathbf{D}_{\alpha\beta} \le \epsilon
\ \text{for all } \alpha,\beta \in \sigma
\right\}.
\]

Thus:
\begin{itemize}
    \item A vertex $\{\alpha\}$ is always included.
    \item An edge $\{\alpha,\beta\}$ is included if $\mathbf{D}_{\alpha\beta} \le \epsilon$.
    \item A $p$-simplex arises whenever all pairwise distances among $p+1$ vertices are at most $\epsilon$.
\end{itemize}

The Vietoris--Rips complex depends only on pairwise distances and provides a combinatorial construction of higher-order structure directly from a distance matrix.

\subsection*{Filtration of Vietoris--Rips complexes}

Given a distance matrix $\mathbf{D}$, the Vietoris--Rips construction admits a natural filtration obtained by varying the threshold parameter $\epsilon$.  
Specifically, one may consider the family of complexes
\[
\mathrm{VR}_{\epsilon}(\mathbf{D}),
\qquad \epsilon \ge 0,
\]
which satisfies
\[
\mathrm{VR}_{\epsilon}(\mathbf{D})
\subseteq
\mathrm{VR}_{\epsilon'}(\mathbf{D})
\quad
\text{whenever } \epsilon \le \epsilon'.
\]
Since the vertex set $S$ is finite, the distance matrix $\mathbf{D}$ contains only finitely many distinct pairwise distances.  
Consequently, new simplices can appear only when $\epsilon$ crosses one of these distance values.  
The Vietoris--Rips construction therefore induces a finite canonical filtration, in which simplices are added at discrete critical thresholds determined entirely by $\mathbf{D}$. This canonical filtration describes how the combinatorial structure evolves as increasingly longer-range interactions are permitted.

In practice, however, it is often unnecessary to use all critical distance values.  
In this work, we utilize a coarse, chemically motivated filtration by selecting a small number of fixed cutoff distances, namely $2\,\text{\AA}$, $3\,\text{\AA}$, and $4\,\text{\AA}$.  
These thresholds correspond to short range (covalent) and long range (non-covalent) interaction ranges and provide a principled multiscale representation suited to molecules.

\section*{Forman curvature}\label{appx:C}
Forman introduced a discrete notion of Ricci curvature \cite{forman_bochners_2003} derived from a combinatorial analogue of the classical Bochner–Weitzenb\"ock formula in Riemannian geometry. In the smooth setting, the Bochner–Weitzenb\"ock formula establishes a fundamental relationship between curvature and the Riemannian Laplace operator \cite{berger2003panoramic, najman_modern_2017, weber2017characterizing}. Although originally formulated for \textit{weighted CW cell complexes}, Forman’s definition extends naturally to polyhedral complexes, simplicial complexes, and graphs, all of which are special instances of cell complexes \cite{forman_bochners_2003, weber2016can}. 

In Riemannian geometry, the Bochner–Weitzenb\"ock formula decomposes the Laplace operator acting on $p$-forms as
$$
\square_p = (\nabla_p)^* \nabla_p + F_p,
$$
where $\square_p$ denotes the Laplace operator on $p$-forms over a compact Riemannian manifold $\mathcal{M}$, $\nabla_p$ is the covariant derivative, and $F_p$ is a zeroth-order operator whose value at a point $x \in \mathcal{M}$ depends only on local derivatives of the metric tensor.

Forman demonstrated that an analogous decomposition can be constructed in the discrete setting:
$$
\square_p = B_p + F_p,
$$
where $\square_p$ now represents the \textit{combinatorial Riemann–Laplace operator}, $B_p$ is the \textit{combinatorial Bochner (or rough) Laplacian}, and $F_p$ defines the $p$th \textit{combinatorial curvature operator}. 
For any $p$-cell $\alpha$, Forman defined the curvature function
$$
\mathcal{F}_p(\alpha) = \langle F_p(\alpha), \alpha \rangle.
$$
While $\mathcal{F}_p$ does not admit a direct geometric interpretation in the smooth setting for $p > 1$, the case $p=1$ coincides with Ricci curvature. Thus, the \textit{Forman–Ricci curvature} of a 1-cell (edge) $e$ is given by $\mathcal{F}_1(e)$. This construction applies generally to weighted cell complexes.

\subsection*{Computation on simplicial complexes}
For a weighted simplicial complex $\mathcal{K}$, the combinatorial Bochner–Weitzenb\"ock framework assigns a curvature value $\mathcal{F}_p(\sigma)$ to each $p$-simplex $\sigma \in \mathcal{K}$. In particular, when $p=1$, the quantity $\mathcal{F}_1(\sigma)$ provides a discrete analogue of classical Ricci curvature on edges.

Two $p$-simplices $\sigma_1, \sigma_2 \in \mathcal{K}_p$ are said to be \textit{parallel}, written $\sigma_1 \parallel \sigma_2$, if exactly one of the following conditions holds:
1. There exists a $(p+1)$-simplex $\tau \in \mathcal{K}_{p+1}$ such that $\sigma_1, \sigma_2 \subset \tau$, or  
2. There exists a $(p-1)$-simplex $\eta \in \mathcal{K}_{p-1}$ such that $\eta \subset \sigma_1, \sigma_2$.
That is, two simplices of equal dimension are parallel if they share either a common coface or a common face, but not both. The set of parallel simplices of $\sigma$ is denoted $\mathrm{Parallel}(\sigma)$.

The Forman–Ricci curvature of an edge (1-simplex) $e \in \mathcal{K}_1$ is defined as
\begin{multline*}
\kappa_F(e) =
w_e
\Bigg[
\left(
\sum_{f \in \mathrm{Coface}(e)} \frac{w_e}{w_f}
+
\sum_{v \in \mathrm{Face}(e)} \frac{w_v}{w_e}
\right) \\
-
\sum_{\hat{e} \in \mathrm{Parallel}(e)}
\left|
\sum_{f \in \mathcal{K}_2 : \hat{e}, e \subset f} \frac{\sqrt{w_e w_{\hat{e}}}}{w_f}
-
\sum_{v \in \mathcal{V} : v \subset \hat{e}, e} \frac{w_v}{\sqrt{w_e w_{\hat{e}}}}
\right|
\Bigg],
\end{multline*}
where $w_\sigma$ denotes the weight assigned to simplex $\sigma$.

In the unweighted case, where $w_\sigma = 1$ for all $\sigma \in \mathcal{K}$, the curvature simplifies to a purely combinatorial expression:
$$
\kappa_F^{\#}(e) =
\#\mathrm{Face}(e)
+
\#\mathrm{Coface}(e)
-
\#\mathrm{Parallel}(e).
$$

This formulation has been applied to various types of abstract simplicial complexes, including clique complexes derived from networks \cite{samal_comparative_2018, chatterjee2021detecting} and Vietoris–Rips complexes constructed from point cloud data \cite{wee2021forman}.

More generally, Forman’s framework provides a principled method to define curvature for simplices of arbitrary dimension. For any weighted simplicial complex $\mathcal{K}$, the combinatorial Bochner–Weitzenb\"ock formula defines a curvature function $\mathcal{F}_p(\sigma)$ for each $p$-simplex $\sigma \in \mathcal{K}$. The case $p=1$ recovers the discrete Ricci curvature on edges, while higher-dimensional analogues depend on the weights of faces, cofaces, and parallel simplices. In the unweighted setting, the curvature reduces to the combinatorial form
$$
\mathcal{F}_p^{\#}(\sigma)
=
\#\mathrm{Face}(\sigma)
+
\#\mathrm{Coface}(\sigma)
-
\#\mathrm{Parallel}(\sigma),
$$
which directly generalizes the edge-based Ricci curvature expression to arbitrary simplices. For a comprehensive treatment of Forman curvature on weighted complexes, we refer the reader to Forman’s original work \cite{forman_bochners_2003}.

\section*{Comparison of HSMP with Hierarchical Message-Passing GNNs}\label{appx:D0}
\begin{figure}[htp!]
    \centering
    \includegraphics[width=0.5\textwidth]{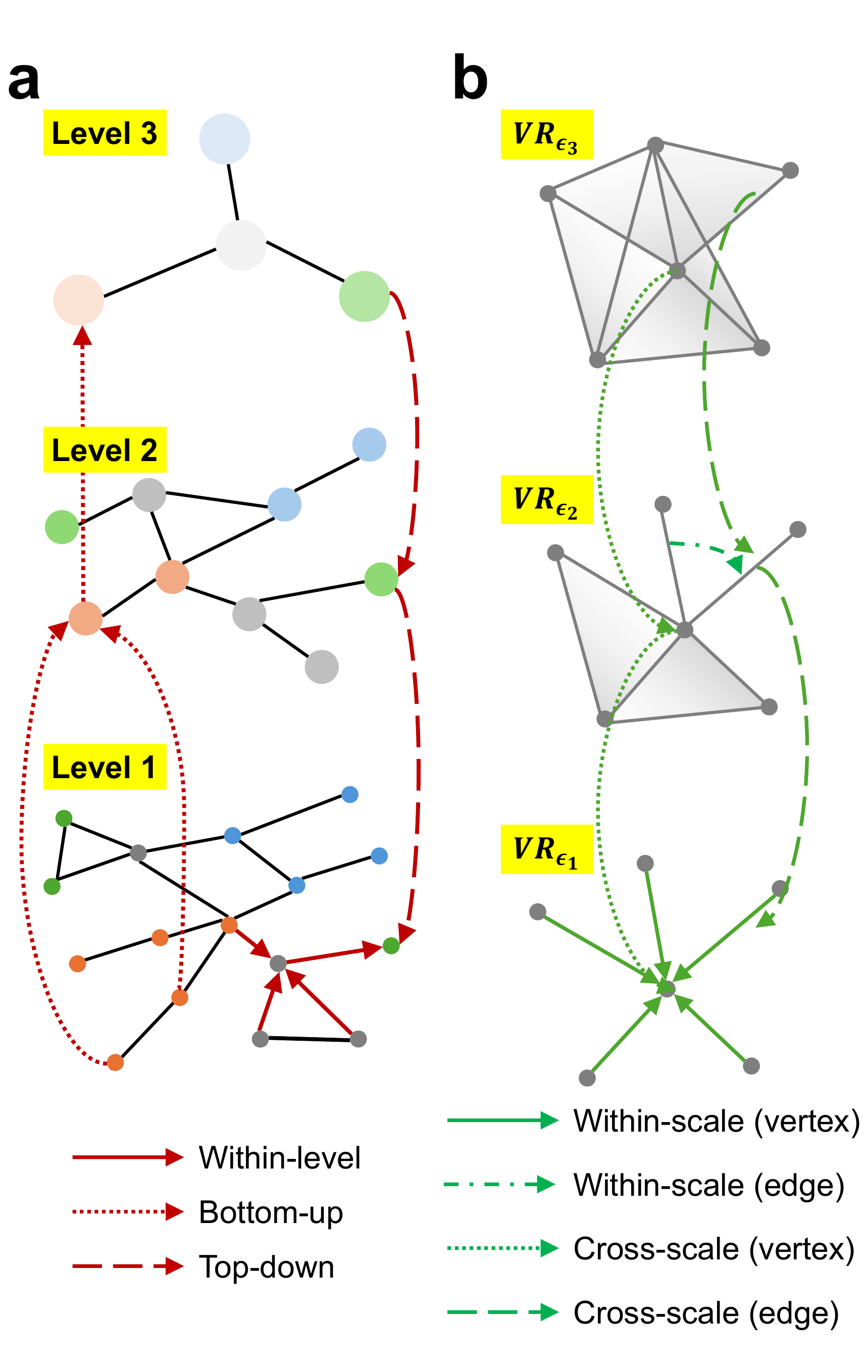} 
    \caption{\textbf{Comparison of information flow in Hierarchical MPGNNs and HSMP.} 
    (a) Hierarchical Message-Passing GNN. The hierarchy is constructed by clustering nodes into disjoint sets of super-nodes across Level 1 to Level 3. Information is propagated via bottom-up (dotted red), top-down (dashed red), and within-level message passing. 
    (b) Hierarchical Simplicial Message Passing (HSMP). The hierarchy is defined by a nested Vietoris--Rips filtration ($VR_{\epsilon_1} \subset VR_{\epsilon_2} \subset VR_{\epsilon_3}$) where the node set remains identical across scales. Information flows within-scale (solid green) and across scales (dotted/dashed green) via a cross-scale refinement (CSR) mechanism, where higher-order topological features at coarser scales modulate representations at finer scales.}
    \label{fig:HSMPComparisonSI}
\end{figure}

Hierarchical Message-Passing Graph Neural Networks have been previously introduced  \cite{zhong_hierarchical_2023} to address two limitations of flat message-passing GNNs, namely the inability to capture long-range interactions and the failure to encode multi-resolution graph semantics. 
In such a framework, nodes are organized into a multi-level hierarchy of super-graphs, where each super-node at level $t$ represents a cluster of nodes from level $t-1$. Super-nodes are typically constructed using community detection algorithms such as Louvain \cite{blondel_fast_2008}. Message passing then proceeds via three propagation schemes to aggregate information across hierarchical levels, namely bottom-up, within-level, and top-down.
Below, we discuss how HSMP fundamentally differs from hierarchical message-passing GNNs in terms of motivation and underlying mechanisms. An illustration comparing the two approaches is provided in Figure~\ref{fig:HSMPComparisonSI}.\\

\paragraph*{Graphs vs. Simplicial Complexes}
Hierarchical MPGNNs operate exclusively on graphs, where message passing is restricted to pairwise interactions along edges. HSMP, by contrast, operates on simplicial complexes, where simplices of dimension $k$ encode simultaneous interactions among $k+1$ atoms. This places HSMP within the framework of topological deep learning, enabling it to explicitly represent many-body interactions that are inaccessible to graph-based approaches.\\

\paragraph*{Absence of Super-nodes Across Levels}
In hierarchical MPGNNs, each level of the hierarchy introduces a distinct, smaller set of super-nodes formed by collapsing groups of lower-level nodes. Hence, the node sets at different levels are disjoint. In the context of molecular data, higher-level nodes might lack a direct chemical interpretation. In HSMP, the node set remains identical across all filtration levels. Every node corresponds to a specific atom in the repeating unit. The hierarchy in HSMP encodes spatial scale, specifically, the filtration parameter $\epsilon_i$ controls the range of interactions, progressing from covalent bonds to non-covalent interactions.\\

\paragraph*{Task-specific Interpretation of Hierarchy}
The hierarchical structure in hierarchical MPGNNs is typically derived from community detection on the graph, which is often task-agnostic and may lack inherent physical meaning. In HSMP, the hierarchy is defined by a Vietoris--Rips filtration over a distance matrix, where each level corresponds to a chemically meaningful interaction range. This ensures that the hierarchy directly reflects the spatial structure of the polymer.\\

\paragraph*{Cross-scale Information Flow}
Hierarchical MPGNNs propagate information across levels through explicit bottom-up and top-down message passing between the original nodes and their associated super-nodes. In HSMP, information flows across scales through the Cross-Scale Refinement (CSR) module, which uses the representation of a simplex at a coarser scale to modulate its representation at a finer scale. Since the nested structure of Vietoris--Rips complexes guarantees that every simplex present at $\epsilon_i$ also exists at $\epsilon_{i+1}$, this cross-scale transfer is well-defined for every simplex without requiring a clustering step.\\

\paragraph*{Compatibility with Molecular Pretraining}
A practical implication of the design choices behind HSMP is that at the finest filtration level ($\epsilon_1 = 2$ \AA), the simplicial complex reduces to the covalent bond graph of the polymer. Node representations at this scale are enriched by higher-order and multi-scale information propagated from coarser levels via CSR, and the resulting graph is structurally compatible with standard molecular graph pretraining frameworks. HSMP can therefore serve as a direct replacement for graph-based self-supervised pretraining, as readout is performed exclusively at the molecular graph level. In contrast, frameworks such as hierarchical MPGNNs require pooling across multiple levels with distinct, disjoint node sets.

\section*{Baseline Model Configurations}\label{appx:D}
All baseline models were evaluated using the same predefined five-fold cross-validation splits as Periodic-TDL. For each fold, 80\% of the data was used for training and 20\% for testing, with 20\% of the training portion reserved for validation. Validation RMSE was used for model selection. Unless otherwise specified, official pretrained weights and default hyperparameters from the original implementations were used without additional tuning or architectural modification.\\

\paragraph*{Morgan (NN).}
Extended-connectivity circular fingerprints (ECFP4; radius 2, 1024 bits) were computed using RDKit from canonical SMILES strings. Bit-based fingerprints were used. The resulting vectors were input to a three-layer multilayer perceptron (hidden dimension 1024, ReLU activation, dropout 0.2) with a final linear regression layer. Models were trained using AdamW with learning rate $3\times10^{-4}$ and weight decay 0.05 for up to 100 epochs. No task-specific tuning was performed.\\

\paragraph*{polyBERT.}
The pretrained polyBERT transformer released by the original authors was finetuned for each downstream task. Canonical SMILES strings were tokenized using the official tokenizer, with polymer anchor symbols normalized to match the pretraining convention. The representation corresponding to the \texttt{[CLS]} token was used as input to a regression head. Finetuning followed a two-stage protocol in which the head was trained prior to joint optimization with the backbone using separate learning rates. Targets were standardized using statistics computed from the training split and inverse-transformed for evaluation.\\

\paragraph*{TransPolymer.}
The pretrained TransPolymer checkpoint provided in the official repository was finetuned using the base configuration specified by the authors. No architectural modifications were introduced. Finetuning was performed under the same cross-validation splits as Periodic-TDL, with early stopping based on validation performance.\\

\paragraph*{polyGNN.}
The polyGNN architecture was implemented using the official library. Polymer graphs were constructed using the monocycle featurization scheme provided by the original implementation. A single-task model was trained independently for each property using the default feature configuration. Training was performed using mean squared error loss without ensembling or hyperparameter optimization.\\

\paragraph*{MolCLR (GCN and GIN).}
Pretrained MolCLR encoders with both GCN and GIN backbones were finetuned for each downstream property without architectural modification. The original configuration provided in the repository was used for optimization and scheduling. Both backbone variants were trained independently under the same data splits and evaluation protocol.\\

\paragraph*{TransChem.}
The TransChem architecture was initialized from the pretrained checkpoint released by the authors. RDKit descriptors were computed according to the feature specification described in the original Supplementary Information and combined with sequence representations as defined in the official implementation. Finetuning followed the base configuration provided in the repository without additional hyperparameter tuning.\\

\paragraph*{MMPolymer.}
The MMPolymer model was initialized from the pretrained weights released by the original authors and finetuned using the default hyperparameter configuration. No architectural modifications or additional tuning were introduced. Evaluation was conducted under the same cross-validation splits as Periodic-TDL.

\section*{Ablation Studies}
\subsection*{Effect of periodic representation and hierarchical message passing}

To evaluate the effect of the periodic Rips complex representation and hierarchical message passing, we performed an ablation study where these components are selectively removed. 
The periodic representation treats the polymer as an extended system utilizing a periodic distance matrix, thereby encoding interactions between atoms belonging to adjacent repeating units. 
In contrast, the non-periodic representation restricts all interactions to atoms within a single monomer.
Hierarchical message passing includes multi-scale propagation of information across filtration levels, where features are refined from coarser to finer scales via cross-scale refinement modules. 
Within each filtration level, message passing is first performed over edges using upper adjacencies and co-boundary adjacencies, followed by node-level message passing informed by the updated edge representations. 
In the non-hierarchical setting, message passing is restricted to within-scale interactions only, without any cross-scale refinement. 
Specifically, edge--edge and node--node message passing are performed independently at each cutoff, and the resulting features are pooled and concatenated to obtain the final polymer representation. 

All results for this ablation study are summarized in Table~\ref{tab:ablation1}, which reports the RMSE and $R^2$ values for three models across all target properties. 
These models are constructed from the HSMP architecture with selective removal of components: 
(i) the full model with both periodic representation and hierarchical message passing, 
(ii) the model without periodic representation but with hierarchical message passing, and
(iii) the model with periodic representation but without hierarchical message passing.
All three models were trained under identical conditions to ensure a fair comparison. 
In particular, we restricted all models to single-head simplicial message passing, removed any effects of pretraining, and keep all training hyperparameters fixed. 
This controlled setup isolates the impact of periodicity and hierarchical propagation on downstream performance.

\subsection*{Effect of multi-head message passing, pretraining, and model capacity}
To evaluate the effect of multi-head simplicial message passing, pretraining, and model capacity, we performed an ablation study where these components are selectively varied.
Multi-head message passing decomposes feature propagation into multiple simplicial message passing heads.
In contrast, the single-head model restricts message passing to a single channel, limiting the diversity of learned representations.
Pretraining initializes the model using a large corpus of polymer structures, allowing it to learn transferable chemical and structural features prior to downstream fine-tuning.
In the absence of pretraining, model parameters are randomly initialized and learned solely from the supervised training data.
Model capacity is controlled through the hidden dimension, with reduced-capacity models employing smaller hidden dimensions.

All configurations considered in this study are summarized in Table~\ref{tab:hsmp_variants}, which reports architectural details including hidden dimension, number of heads, pretraining data size, and the number of trainable parameters.
The evaluated models correspond to variations of the HSMP architecture with controlled modifications:
(i) MH-Base (Pretrained), the full model with multi-head message passing and pretraining,
(ii) MH-Base (Random), which removes pretraining while retaining the same architecture,
(iii) SH-Base (Pretrained), which replaces multi-head message passing with a single-head version at fixed capacity,
(iv) MH-Small (Pretrained), which reduces model capacity while retaining multi-head message passing and pretraining, and
(v) SH-Small (Pretrained), which combines reduced capacity with single-head message passing.

All models were trained under identical conditions to ensure a fair comparison.
In particular, we keep the periodic representation and hierarchical message passing fixed across all models, and maintain identical training hyperparameters.
This controlled setup isolates the impact of multi-head message passing, pretraining, and model capacity on downstream performance.
The corresponding performance improvements relative to the best baseline model for each dataset and metric are reported in Table~\ref{tab:hsmp_vs_best_baseline}, enabling a direct comparison of how each design choice across model configurations contributes to predictive performance.

\section*{Synthesis and Characterization of Poly(2-(acryloyloxy)ethyl benzoate), Poly(2-(methacryloyloxy)ethyl benzoate), and Poly(2-acrylamidoethyl benzoate)}

\paragraph*{\textbf{Materials}}
Benzoic acid (98\%, Sigma Aldrich, USA), 2-hydroxyethyl methacrylate (HEMA) ($>$95\%, Tokyo Chemical Industry (TCI), Japan), 2-hydroxyethyl acrylate (HEA) ($>$95\%, Tokyo Chemical Industry (TCI), Japan), 2-hydroxyethyl acrylamide (HEAm) ($>$95\%, Sigma Aldrich, Japan), 4-dimethylaminopyridine (DMAP) ($\geq$99\%, Sigma Aldrich), 1-(3-dimethylaminopropyl)-3-ethylcarbodiimide hydrochloride (EDC$\cdot$HCl) ($>$98\%, TCI), N,N$'$-dicyclohexylcarbodiimide (DCC) (99\%, Sigma Aldrich), hydrogen peroxide solution (35\% in water) (TCI), 2,2$'$-azoisobutyronitrile (AIBN) (95\%, Fujifilm Wako Pure Chemical, Japan), dichloromethane (DCM) (99.8\%, Fisher Chemicals, USA), anhydrous 1,4-dioxane (99.8\%, Sigma Aldrich), and N,N-dimethylformamide (DMF) ($>$99.5\%, TCI) were used as received.\\

\paragraph*{\textbf{Synthesis of 2-(Methacryloyloxy)ethyl benzoate}}
A reaction mixture of 10.00 g ($7.68 \times 10^{-2}$ mol) 2-hydroxyethyl methacrylate, 14.73 g ($7.68 \times 10^{-2}$ mol) 1-ethyl-3-(3-dimethylaminopropyl)carbodiimide hydrochloride (EDC$\cdot$HCl), and 1.70 g ($1.71 \times 10^{-2}$ mol) 4-(dimethylamino)pyridine (DMAP) was dissolved in 100 mL dichloromethane. The mixture was stirred for 30 min at room temperature, after which 8.53 g ($6.98 \times 10^{-2}$ mol) benzoic acid was added. The reaction was stirred for 16 h at room temperature. The mixture was then washed three times with 150 mL H$_2$O. The organic phase was washed once more with 100 mL brine and dried over MgSO$_4$. The solvent was removed to obtain a pale yellow liquid. Yield: 9.53 g (40.6\%). $^1$H NMR (CDCl$_3$, 400 MHz): $\delta$ 8.04 (d, 2H), 7.55 (t, 1H), 7.43 (t, 2H), 6.14 (s, 1H), 5.57 (s, 3H), 4.53 (m, 4H), 1.94 (s, 3H).\\

\paragraph*{\textbf{Synthesis of 2-(Acryloyloxy)ethyl benzoate}}
A reaction mixture of 10.00 g ($8.61 \times 10^{-2}$ mol) 2-hydroxyethyl acrylate, 9.56 g ($7.86 \times 10^{-2}$ mol) EDC$\cdot$HCl, and 1.91 g ($1.57 \times 10^{-2}$ mol) DMAP was dissolved in 100 mL dichloromethane. The mixture was stirred for 30 min at room temperature, followed by the addition of 9.56 g ($7.82 \times 10^{-2}$ mol) benzoic acid. The reaction was stirred for 16 h at room temperature. The mixture was then washed three times with 150 mL H$_2$O. The organic phase was washed once more with 100 mL brine and dried over MgSO$_4$. The solvent was removed to obtain a pale yellow liquid. Yield: 10.61 g (48.2\%). $^1$H NMR (CDCl$_3$, 400 MHz): $\delta$ 7.99 (d, 2H), 7.48 (t, 1H), 7.36 (t, 2H), 6.36 (d, 1H), 6.09 (q, 1H), 5.77 (d, 1H), 4.46 (m, 4H).\\

\paragraph*{\textbf{Synthesis of 2-acrylamidoethyl benzoate}}
A reaction mixture of 10.00 g ($8.68 \times 10^{-2}$ mol) 2-hydroxyethyl acrylamide, 9.64 g ($8.68 \times 10^{-2}$ mol) EDC$\cdot$HCl, and 1.93 g ($1.58 \times 10^{-2}$ mol) DMAP was dissolved in 100 mL dichloromethane. The mixture was stirred for 30 min at room temperature, after which 9.64 g ($7.90 \times 10^{-2}$ mol) benzoic acid was added. The reaction was stirred for 16 h at room temperature. The mixture was then washed three times with 150 mL H$_2$O. The organic phase was washed once more with 100 mL brine and dried over MgSO$_4$. The solvent was removed to obtain a white solid. Yield: 9.82 g (44.8\%). $^1$H NMR (CDCl$_3$, 400 MHz): $\delta$ 8.06 (d, 2H), 7.60 (t, 1H), 7.47 (t, 2H), 6.31 (d, 1H), 6.14 (q, 1H), 5.67 (d, 1H), 4.47 (t, 2H), 3.76 (q, 2H), 1.79 (br, 1H).\\

\paragraph*{\textbf{Synthesis of polymers}}
Homopolymers were synthesized via radical polymerization. A mixture of monomer (typically 2 g), AIBN (typically 0.014 g), and 4.5 g DMF in a Schlenk flask was heated at 70 $^\circ$C for 20 h under an argon atmosphere with magnetic stirring. The polymer was purified via reprecipitation in diethyl ether (non-solvent) three times and dried overnight under vacuum. The polymerization results are given in Table \ref{tab:polymerization}.\\

\paragraph*{\textbf{Gel permeation chromatography (GPC)}}
GPC analysis was performed using DMF as the eluent on a Shimadzu i-Series Plus liquid chromatograph LC-2030C Plus (Kyoto, Japan) equipped with two Shodex LF-804 columns (300 $\times$ 8.0 mm; bead size = 6 $\mu$m; pore size = 3000~\AA) and one Shodex KD-802 column (300 $\times$ 8.0 mm; bead size = 6 $\mu$m; pore size = 150~\AA). The DMF eluent contained 10 mM LiBr, and the flow rate was 0.34 mL/min (40 $^\circ$C). Sample detection was conducted using a Shimadzu differential refractometer detector RID-20A, and the column system was calibrated with standard poly(methyl methacrylate)s (PMMAs).\\

\paragraph*{\textbf{NMR spectroscopy}}
$^1$H NMR and $^{13}$C NMR spectra were recorded on a Bruker BBFO400 spectrometer (400 MHz) at ambient temperature. Chloroform-$d$ (CDCl$_3$) (Cambridge Isotope Laboratories, USA) or DMSO-$d_6$ (Cambridge Isotope Laboratories) was used as the NMR solvent. Chemical shifts were calibrated using the residual undeuterated solvent or tetramethylsilane (TMS) as an internal standard. Monomer conversions in the polymerization were determined using $^1$H NMR from the decay in the peak areas of the vinyl groups of the monomers.\\

\paragraph*{\textbf{Differential scanning calorimetry (DSC)}}
The glass transition temperatures ($T_g$) of the three polymers were obtained using differential scanning calorimetry (DSC), specifically a QSeries DSC Q50 model device (TA Instruments, New Castle, USA). DSC analysis was conducted using aluminium sample pans under nitrogen flow at 50 mL/min. Samples were cooled to $-60$ $^\circ$C and heated to 200 $^\circ$C at a heating rate of 10 $^\circ$C/min.\\


\clearpage
\section*{Supplementary Figures}\label{appx:F}
\begin{figure}[htp!]
\centering
\includegraphics[width=\textwidth]{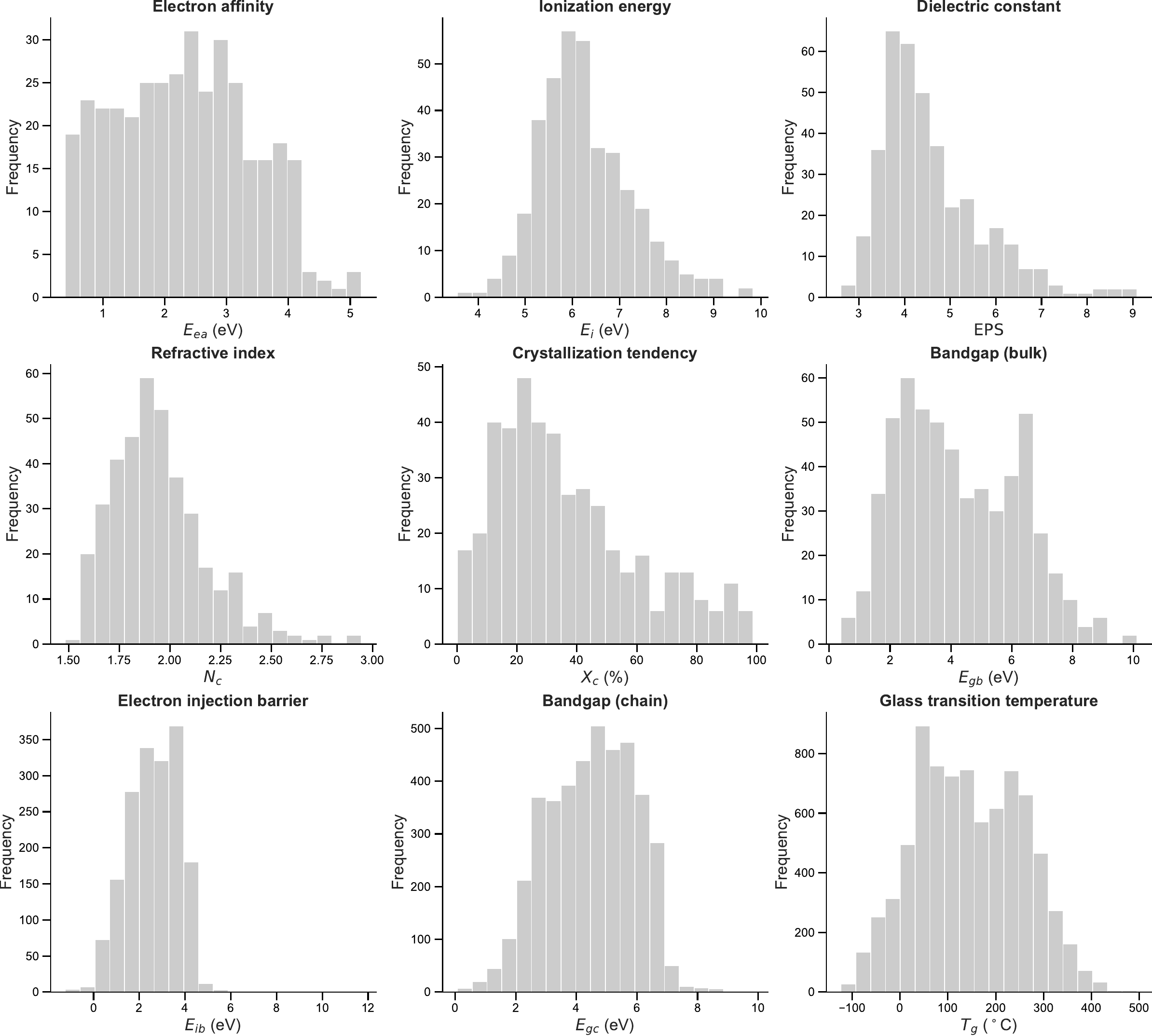}
\caption{\textbf{Distributions of true target values for the nine downstream polymer property prediction tasks evaluated in this work.} Histograms are shown for electronic ($E_{ea}$, $E_{i}$, $E_{gc}$, $E_{ib}$, $E_{gb}$), optical (EPS, $N_{c}$), physical ($X_{c}$), and thermal ($T_{g}$) properties. The horizontal axis in each panel corresponds to the property value (with units indicated), and the vertical axis corresponds to the frequency of occurrences within the corresponding bin.%
}
\label{fig:appendix_property_histograms}
\end{figure}
\begin{figure}[htp!]
\centering
\includegraphics[width=\textwidth]{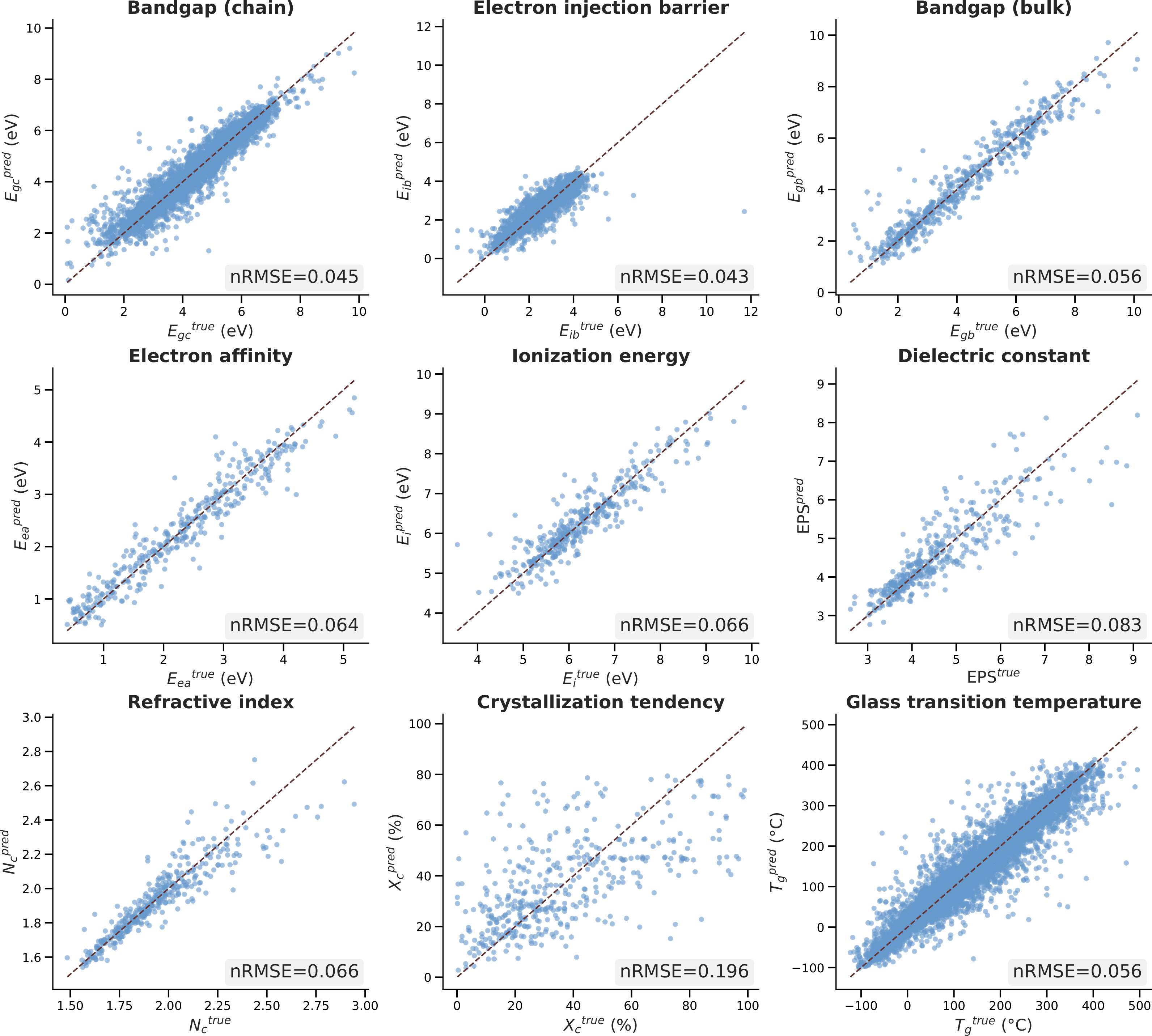}
\caption{\textbf{Parity plots comparing true versus predicted values from Periodic-TDL for all test samples across five cross-validation folds and nine downstream polymer property tasks.} Each panel corresponds to one property target. Points are pooled across test folds, and the dashed line indicates a perfect agreement between predicted and true values. Normalized RMSE values are shown in the individual plot panels.}
\label{fig:appendix_parity_plots}
\end{figure}
\begin{figure}[htp!]
    \centering
    \includegraphics[width=\textwidth]{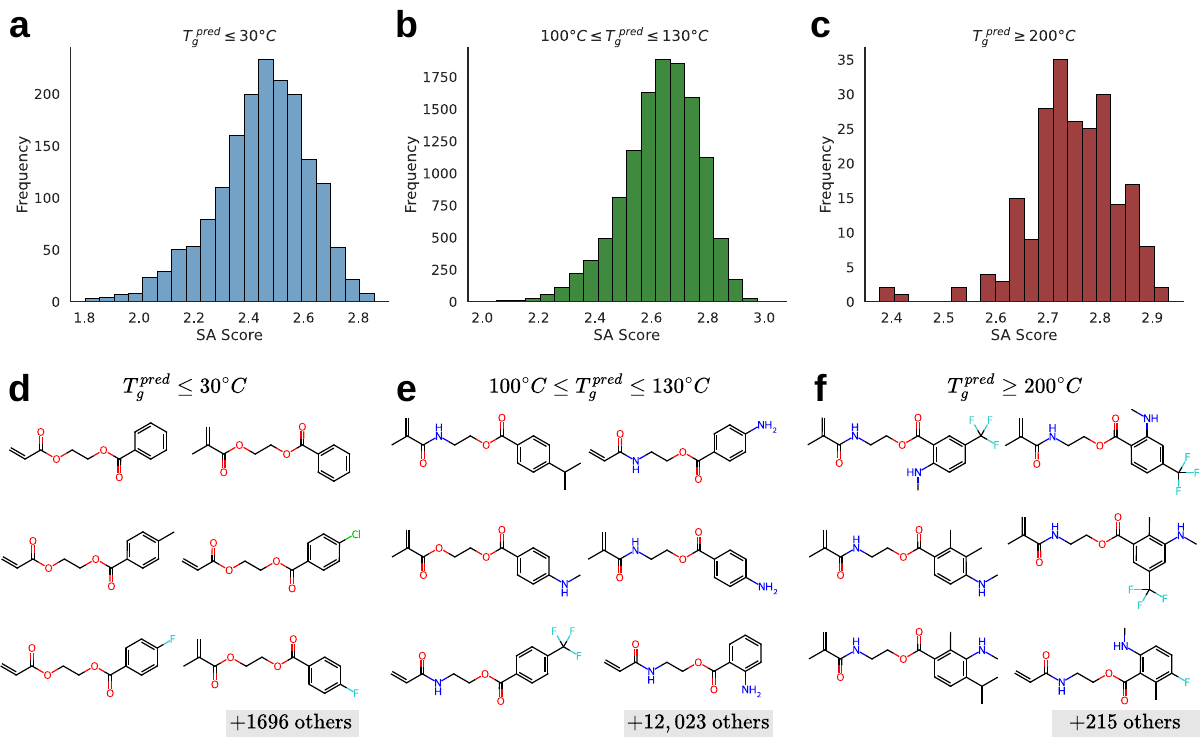}
    \caption{\textbf{Synthetic accessibility of monomers corresponding to polymers in different predicted glass transition temperature ranges.}
    \textbf{(a--c)} Distributions of Schuffenhauer synthetic accessibility (SA) scores for monomers corresponding to polymers with low ($T_g^{\mathrm{pred}} \leq 30\,^\circ\mathrm{C}$), medium ($100\,^\circ\mathrm{C} \leq T_g^{\mathrm{pred}} \leq 130\,^\circ\mathrm{C}$), and high ($T_g^{\mathrm{pred}} \geq 200\,^\circ\mathrm{C}$) glass transition temperature ranges, respectively.
    \textbf{(d--f)} Representative monomers from each $T_g^{\mathrm{pred}}$ range, showing the six monomers with the lowest SA scores within each range.
    These ranges contain 1,702, 12,029, and 221 polymers, respectively.
    Higher $T_g^{\mathrm{pred}}$ ranges are associated with bulkier, more polar and hydrogen-bonding prone polymers.}
    \label{fig:sascore}
\end{figure}

\clearpage
\section*{Supplementary Tables}\label{appx:E}
\begin{table}[htp!]
\centering
\renewcommand{\arraystretch}{2.0}
\setlength{\tabcolsep}{6pt}
\scriptsize
\caption{\textbf{Summary statistics of the nine downstream polymer property prediction datasets.}
For each target, we report the number of samples ($n$), mean, standard deviation (SD), minimum, maximum, and empirical range ($\max - \min$).
These statistics are used to compare RMSE with the standard deviation of each dataset and to compute normalized RMSE, defined as $\mathrm{RMSE}/(\max(y)-\min(y))$, as reported in the Results.
Electronic ($E_{gc}$, $E_{ib}$, $E_{gb}$, $E_{ea}$, $E_i$), optical (EPS, $N_c$), and physical ($X_c$) properties are derived from density functional theory calculations, whereas $T_g$ corresponds to experimentally measured values.}
\label{tab:appendix_table1_dataset_stats}
\begin{tabular}{lllc ccccccc}
\toprule
\textbf{Property} & \textbf{Property name} & \textbf{Class} & \textbf{Unit} &
\textbf{$n$} & \textbf{Mean} & \textbf{SD} & \textbf{Min} & \textbf{Max} & \textbf{Range} \\
\midrule
$E_{ea}$ & Electron affinity & Electronic & eV &
368 & 2.307 & 1.089 & 0.394 & 5.178 & 4.785 \\

$E_i$ & Ionization energy & Electronic & eV &
370 & 6.268 & 0.991 & 3.558 & 9.839 & 6.281 \\

EPS & Dielectric constant & Optical & -- &
382 & 4.586 & 1.119 & 2.610 & 9.090 & 6.480 \\

$N_c$ & Refractive index & Optical & -- &
382 & 1.947 & 0.240 & 1.484 & 2.945 & 1.460 \\

$X_c$ & Crystallization tendency & Physical & \% &
430 & 36.76 & 23.71 & 0.13 & 98.81 & 98.68 \\

$E_{gb}$ & Bandgap (bulk) & Electronic & eV &
561 & 4.242 & 1.954 & 0.391 & 10.114 & 9.723 \\

$E_{ib}$ & Electron injection barrier & Electronic & eV &
1744 & 2.636 & 1.107 & -1.234 & 11.703 & 12.937 \\

$E_{gc}$ & Bandgap (chain) & Electronic & eV &
4124 & 4.488 & 1.442 & 0.071 & 9.835 & 9.764 \\

$T_g$ & Glass transition temperature & Thermal & $^\circ$C &
7957 & 145.87 & 110.50 & -123.00 & 495.00 & 618.00 \\
\bottomrule
\end{tabular}
\end{table}
\begin{table}[htp!]
\centering
\renewcommand{\arraystretch}{2.0}
\setlength{\tabcolsep}{6pt}
\scriptsize
\caption{\textbf{Test coefficient of determination ($R^2$) of Periodic-TDL and baseline models across nine polymer property prediction tasks.}
Values are reported as mean $\pm$ standard deviation across five test folds, with all models evaluated using identical data splits.
Periodic-TDL (+Chem) combines Periodic-TDL with global chemical descriptor-based predictors through linear stacking, and Periodic-TDL (+Chem, +RC) further incorporates a local residual correction based on polymer similarity.
Higher values indicate better performance.
The best and second-best mean values for each target are shown in \textbf{bold} and \underline{underlined}, respectively.}
\label{tab:r2_comparison}
\resizebox{\textwidth}{!}{%
\begin{tabular}{lccccccccc}
\toprule
\textbf{Model} &
\textbf{$E_{ea}$} &
\textbf{$E_i$} &
\textbf{$EPS$} &
\textbf{$N_c$} &
\textbf{$X_c$} &
\textbf{$E_{gb}$} &
\textbf{$E_{ib}$} &
\textbf{$E_{gc}$} &
\textbf{$T_g$} \\
\midrule

Morgan (NN) &
0.840 $\pm$ 0.009 &
0.439 $\pm$ 0.115 &
0.552 $\pm$ 0.092 &
0.497 $\pm$ 0.122 &
0.320 $\pm$ 0.095 &
0.802 $\pm$ 0.026 &
0.662 $\pm$ 0.048 &
0.849 $\pm$ 0.010 &
0.882 $\pm$ 0.010 \\

polyBERT &
0.904 $\pm$ 0.015 &
0.794 $\pm$ 0.024 &
0.674 $\pm$ 0.089 &
0.768 $\pm$ 0.064 &
0.298 $\pm$ 0.087 &
0.895 $\pm$ 0.018 &
0.730 $\pm$ 0.048 &
0.887 $\pm$ 0.012 &
0.889 $\pm$ 0.011 \\

TransPolymer &
0.901 $\pm$ 0.017 &
0.783 $\pm$ 0.028 &
0.652 $\pm$ 0.080 &
0.756 $\pm$ 0.058 &
0.338 $\pm$ 0.073 &
0.899 $\pm$ 0.032 &
0.715 $\pm$ 0.053 &
0.896 $\pm$ 0.016 &
0.894 $\pm$ 0.014 \\

polyGNN &
0.917 $\pm$ 0.018 &
0.621 $\pm$ 0.188 &
0.654 $\pm$ 0.157 &
0.778 $\pm$ 0.074 &
0.185 $\pm$ 0.142 &
0.902 $\pm$ 0.027 &
0.725 $\pm$ 0.048 &
0.867 $\pm$ 0.020 &
0.846 $\pm$ 0.005 \\

MolCLR (GCN) &
0.867 $\pm$ 0.015 &
0.693 $\pm$ 0.097 &
0.675 $\pm$ 0.087 &
0.745 $\pm$ 0.076 &
0.238 $\pm$ 0.089 &
0.875 $\pm$ 0.021 &
0.725 $\pm$ 0.053 &
0.850 $\pm$ 0.009 &
0.835 $\pm$ 0.014 \\

MolCLR (GIN) &
0.897 $\pm$ 0.021 &
0.644 $\pm$ 0.189 &
0.664 $\pm$ 0.063 &
0.735 $\pm$ 0.077 &
0.231 $\pm$ 0.089 &
0.894 $\pm$ 0.016 &
0.747 $\pm$ 0.043 &
0.886 $\pm$ 0.009 &
0.859 $\pm$ 0.008 \\

TransChem &
0.876 $\pm$ 0.038 &
0.732 $\pm$ 0.120 &
0.637 $\pm$ 0.100 &
0.769 $\pm$ 0.049 &
0.309 $\pm$ 0.039 &
0.883 $\pm$ 0.034 &
0.660 $\pm$ 0.104 &
0.887 $\pm$ 0.017 &
0.895 $\pm$ 0.010 \\

MMPolymer &
0.899 $\pm$ 0.011 &
0.743 $\pm$ 0.066 &
0.601 $\pm$ 0.112 &
0.744 $\pm$ 0.054 &
0.325 $\pm$ 0.136 &
0.903 $\pm$ 0.020 &
0.699 $\pm$ 0.027 &
0.903 $\pm$ 0.010 &
0.895 $\pm$ 0.012 \\

PerioGT &
0.929 $\pm$ 0.016 &
0.792 $\pm$ 0.062 &
0.741 $\pm$ 0.084 &
0.821 $\pm$ 0.067 &
\second{0.421 $\pm$ 0.080} &
0.909 $\pm$ 0.023 &
0.755 $\pm$ 0.052 &
\second{0.909 $\pm$ 0.010} &
\second{0.915 $\pm$ 0.008} \\

\midrule

Periodic-TDL &
\best{0.934 $\pm$ 0.013} &
\best{0.839 $\pm$ 0.018} &
\best{0.787 $\pm$ 0.033} &
\best{0.856 $\pm$ 0.032} &
\best{0.423 $\pm$ 0.102} &
\second{0.921 $\pm$ 0.012} &
\second{0.758 $\pm$ 0.045} &
\second{0.909 $\pm$ 0.011} &
0.904 $\pm$ 0.010 \\

Periodic-TDL (+Chem) &
\second{0.931 $\pm$ 0.011} &
0.824 $\pm$ 0.010 &
0.783 $\pm$ 0.032 &
\second{0.849 $\pm$ 0.035} &
0.414 $\pm$ 0.079 &
\second{0.921 $\pm$ 0.012} &
\best{0.772 $\pm$ 0.046} &
0.908 $\pm$ 0.009 &
0.912 $\pm$ 0.010 \\

Periodic-TDL (+Chem, +RC) &
\second{0.931 $\pm$ 0.011} &
\second{0.826 $\pm$ 0.005} &
\second{0.784 $\pm$ 0.048} &
0.845 $\pm$ 0.052 &
0.394 $\pm$ 0.078 &
\best{0.923 $\pm$ 0.009} &
0.743 $\pm$ 0.046 &
\best{0.911 $\pm$ 0.008} &
\best{0.916 $\pm$ 0.009} \\

\bottomrule
\end{tabular}
}
\end{table}
\begin{table}[htp!]
\centering
\renewcommand{\arraystretch}{2.0}
\setlength{\tabcolsep}{6pt}
\scriptsize
\caption{\textbf{Effect of periodic representation and hierarchical message passing.}
RMSE ($\downarrow$) and $R^2$ ($\uparrow$) values across all target properties for three models defined by the inclusion of periodic representation and hierarchical message passing: periodic + hierarchical, non-periodic + hierarchical, and periodic + non-hierarchical. 
All models are trained under identical settings with single-head simplicial message passing and without pretraining.}
\label{tab:ablation1}
\begin{tabular}{lcccccc}
\toprule
&
\multicolumn{2}{c}{\textbf{Periodic + Hierarchical}} &
\multicolumn{2}{c}{\textbf{Non-periodic + Hierarchical}} &
\multicolumn{2}{c}{\textbf{Periodic + Non-hierarchical}} \\
\cmidrule(lr){2-3}
\cmidrule(lr){4-5}
\cmidrule(lr){6-7}
\textbf{Property} & RMSE $\downarrow$ & $R^2$ $\uparrow$
& RMSE $\downarrow$ & $R^2$ $\uparrow$
& RMSE $\downarrow$ & $R^2$ $\uparrow$ \\
\midrule
$E_{ea}$ & \best{0.281 $\pm$ 0.034} & \best{0.9308 $\pm$ 0.0156} & 0.299 $\pm$ 0.026 & 0.9224 $\pm$ 0.0150 & 0.305 $\pm$ 0.030 & 0.9190 $\pm$ 0.0186 \\
$E_i$    & 0.464 $\pm$ 0.082 & 0.7756 $\pm$ 0.0290 & 0.451 $\pm$ 0.064 & 0.7873 $\pm$ 0.0189 & \best{0.448 $\pm$ 0.080} & \best{0.7908 $\pm$ 0.0251} \\
EPS      & \best{0.580 $\pm$ 0.068} & \best{0.7263 $\pm$ 0.0520} & 0.607 $\pm$ 0.048 & 0.7010 $\pm$ 0.0420 & 0.582 $\pm$ 0.067 & 0.7159 $\pm$ 0.0920 \\
$N_c$    & \best{0.099 $\pm$ 0.006} & \best{0.8223 $\pm$ 0.0421} & 0.110 $\pm$ 0.008 & 0.7814 $\pm$ 0.0483 & 0.100 $\pm$ 0.007 & 0.8189 $\pm$ 0.0472 \\
$X_c$    & \best{18.64 $\pm$ 1.07} & \best{0.3746 $\pm$ 0.0721} & 18.84 $\pm$ 0.82 & 0.3603 $\pm$ 0.0764 & 19.36 $\pm$ 1.15 & 0.3239 $\pm$ 0.0839 \\
$E_{gb}$ & \best{0.549 $\pm$ 0.064} & \best{0.9202 $\pm$ 0.0083} & 0.575 $\pm$ 0.076 & 0.9125 $\pm$ 0.0126 & 0.594 $\pm$ 0.082 & 0.9067 $\pm$ 0.0136 \\
$E_{ib}$ & \best{0.544 $\pm$ 0.062} & \best{0.7563 $\pm$ 0.0400} & 0.560 $\pm$ 0.053 & 0.7428 $\pm$ 0.0315 & 0.551 $\pm$ 0.059 & 0.7509 $\pm$ 0.0365 \\
$E_{gc}$ & \best{0.459 $\pm$ 0.027} & \best{0.8981 $\pm$ 0.0118} & 0.464 $\pm$ 0.026 & 0.8956 $\pm$ 0.0117 & 0.469 $\pm$ 0.025 & 0.8935 $\pm$ 0.0125 \\
$T_g$    & \best{38.11 $\pm$ 1.46} & \best{0.8808 $\pm$ 0.0091} & 38.60 $\pm$ 1.79 & 0.8776 $\pm$ 0.0120 & 38.59 $\pm$ 1.06 & 0.8778 $\pm$ 0.0082 \\
\bottomrule
\end{tabular}
\end{table}
\begin{table}[htp!]
\centering
\renewcommand{\arraystretch}{2.0}
\setlength{\tabcolsep}{8pt}
\scriptsize
\caption{\textbf{Architectural and training details of Periodic-TDL models used in the ablation study.}
We report the hidden dimension, number of attention heads, size of the pretraining dataset, and the number of trainable parameters for each model configuration.}
\label{tab:hsmp_variants}
\begin{tabular}{lcccc}
\toprule
\textbf{Model} &
\textbf{Hidden Dim.} &
\textbf{\# Heads} &
\textbf{Pretraining Size} &
\textbf{\# Parameters} \\
\midrule

MH-Base (Pretrained) &
768 & 12 & $\sim$1M & 35,019,889 \\

MH-Base (Random) &
768 & 12 & 0 & 35,019,889 \\

SH-Base (Pretrained) &
768 & 1 & $\sim$1M & 91,249,205 \\

MH-Small (Pretrained) &
512 & 8 & 250K & 16,792,993 \\

SH-Small (Pretrained) &
512 & 1 & 250K & 40,647,733 \\
\bottomrule
\end{tabular}
\end{table}
\begin{table}[htp!]
\centering
\renewcommand{\arraystretch}{2.0}
\setlength{\tabcolsep}{6pt}
\scriptsize
\caption{\textbf{Relative performance improvement of Periodic-TDL models over best baseline models.}
For each dataset, we report the percentage change in test RMSE and the absolute change in test $R^2$ of different model configurations with respect to the best-performing baseline model for the corresponding metric and dataset.
The final row reports the average improvement across all datasets, enabling an overall comparison across model configurations.}
\label{tab:hsmp_vs_best_baseline}
\resizebox{\textwidth}{!}{%
\begin{tabular}{lcccccccccc}
\toprule
\textbf{Dataset} &
\multicolumn{2}{c}{\textbf{MH-Base (Pretrained)}} &
\multicolumn{2}{c}{\textbf{MH-Base (Random)}} &
\multicolumn{2}{c}{\textbf{SH-Base (Pretrained)}} &
\multicolumn{2}{c}{\textbf{MH-Small (Pretrained)}} &
\multicolumn{2}{c}{\textbf{SH-Small (Pretrained)}} \\
\cmidrule(lr){2-3}
\cmidrule(lr){4-5}
\cmidrule(lr){6-7}
\cmidrule(lr){8-9}
\cmidrule(lr){10-11}
& RMSE (\%) $\downarrow$ & $R^2$ $\uparrow$
& RMSE (\%) $\downarrow$ & $R^2$ $\uparrow$
& RMSE (\%) $\downarrow$ & $R^2$ $\uparrow$
& RMSE (\%) $\downarrow$ & $R^2$ $\uparrow$
& RMSE (\%) $\downarrow$ & $R^2$ $\uparrow$ \\
\midrule

$E_{ea}$ &
$-3.49$ & $+0.005$ &
$+0.08$ & $+0.000$ &
$+5.05$ & $-0.009$ &
$+5.38$ & $-0.008$ &
$+26.93$ & $-0.044$ \\

$E_i$ &
$-11.27$ & $+0.045$ &
$+2.00$ & $-0.004$ &
$-8.07$ & $+0.032$ &
$+4.74$ & $-0.020$ &
$+10.89$ & $-0.047$ \\

EPS &
$-8.02$ & $+0.046$ &
$-0.08$ & $+0.008$ &
$+3.83$ & $-0.019$ &
$+15.39$ & $-0.087$ &
$+9.84$ & $-0.051$ \\

$N_c$ &
$-8.46$ & $+0.035$ &
$+1.15$ & $+0.003$ &
$+4.24$ & $-0.006$ &
$+19.16$ & $-0.067$ &
$+40.53$ & $-0.165$ \\

$X_c$ &
$-0.44$ & $+0.002$ &
$+4.35$ & $-0.049$ &
$+2.08$ & $-0.027$ &
$+10.51$ & $-0.126$ &
$+12.50$ & $-0.158$ \\

$E_{gb}$ &
$-6.55$ & $+0.013$ &
$-2.81$ & $+0.006$ &
$-4.47$ & $+0.009$ &
$+2.16$ & $-0.003$ &
$+6.59$ & $-0.012$ \\

$E_{ib}$ &
$-0.49$ & $+0.002$ &
$+0.87$ & $-0.004$ &
$+3.22$ & $-0.015$ &
$+5.09$ & $-0.025$ &
$-0.15$ & $+0.000$ \\

$E_{gc}$ &
$+0.35$ & $-0.001$ &
$+3.88$ & $-0.007$ &
$+6.99$ & $-0.013$ &
$+9.34$ & $-0.018$ &
$+10.70$ & $-0.020$ \\

$T_g$ &
$+6.25$ & $-0.011$ &
$+8.09$ & $-0.014$ &
$+8.76$ & $-0.015$ &
$+15.58$ & $-0.028$ &
$+26.38$ & $-0.050$ \\

\midrule
\textbf{Average} &
$-3.57$ & $+0.015$ &
$+1.95$ & $-0.007$ &
$+2.40$ & $-0.007$ &
$+9.71$ & $-0.042$ &
$+16.02$ & $-0.061$ \\
\bottomrule
\end{tabular}
}
\end{table}
\begin{table}[htp!]
\centering
\renewcommand{\arraystretch}{2.0}
\setlength{\tabcolsep}{6pt}
\scriptsize
\caption{Summary statistics of predicted glass transition temperature ($T_g^{\mathrm{pred}}$) values for the four systematically substituted polymer families.
Each family contains polymers generated from 12,052 substituted monomers, with substitutions performed on the phenyl ring.
The four families correspond to poly(acrylates) (Ar--Et--A), poly(methacrylates) (Ar--Et--MA), poly(acrylamides) (Ar--Et--AM), and poly(methacrylamides) (Ar--Et--MAM).}
\label{tab:supp_family_tg_stats}
\begin{tabular}{lllc}
\toprule
\textbf{Name} & \textbf{Family} & \textbf{Functional group} & \textbf{$T_g^{\mathrm{pred}}$ ($^\circ$C)} \\
\midrule
Ar--Et--A & Poly(acrylate) & Ester & $71.51 \pm 35.84$ \\
Ar--Et--MA & Poly(methacrylate) & Ester & $86.93 \pm 33.86$ \\
Ar--Et--AM & Poly(acrylamide) & Amide & $128.48 \pm 28.39$ \\
Ar--Et--MAM & Poly(methacrylamide) & Amide & $141.49 \pm 26.59$ \\
\bottomrule
\end{tabular}
\end{table}
\begin{table}[htp!]
\centering
\renewcommand{\arraystretch}{2.0}
\setlength{\tabcolsep}{6pt}
\scriptsize
\caption{\textbf{Individual $T_g$ values for literature-based validation pairs.} For the 8 polymers comprising  the four literature pairs, we report the polymer abbreviation, full polymer name, experimentally reported $T_g$, and  Periodic TDL-predicted $T_g$ (mean $\pm$ SD across ten cross-validation folds). The $\Delta T_g$ values shown in Table 3 are calculated from the individual $T_g$ values for each fold. }
\label{tab:supp_literature_tg_validation}
\begin{tabular}{llcc}
\toprule
\textbf{Abbreviation} & \textbf{Polymer} & \textbf{$T_g^{lit}$ ($^\circ$C)} & \textbf{$T_g^{pred}$ ($^\circ$C)} \\
\midrule
Poly-O & poly[N-[4-(methacryloyloxy)phenyl]-2-(4-methoxyphenyl)acetamide] & 144.8 & $157.9 \pm 7.7$ \\
Poly-A & poly[N-[4-[[(4-methoxyphenyl)acetyl]oxy]phenyl]methacrylamide] & 241.8 & $166.3 \pm 8.2$ \\
PSA & Poly(syringaldehyde acrylate) & 170.0 & $88.0 \pm 10.0$ \\
PSMA & Poly(syringaldehyde methacrylate) & 180.0 & $99.1 \pm 10.6$ \\
PVA & Poly(vanillin acrylate) & 95.0 & $71.2 \pm 11.5$ \\
PVMA & Poly(vanillin methacrylate) & 110.0 & $88.6 \pm 11.3$ \\
PGlc-$\beta$-EAAM & Poly(N-(beta-glycosyloxy)-ethyl acrylamide) & 144.5 & $106.3 \pm 8.3$ \\
PGlc-$\beta$-EMAAM & Poly(N-(beta-glycosyloxy)-ethyl methacrylamide) & 168.0 & $110.5 \pm 7.7$ \\
\bottomrule
\end{tabular}
\end{table}
\begin{table}[htp!]
\centering
\footnotesize
\renewcommand{\arraystretch}{1.3}
\setlength{\tabcolsep}{4pt}
\caption{Polymerization results}
\label{tab:polymerization}
\begin{tabular}{cccccccc}
\toprule
Entry & Monomer & $[\mathrm{M}]_0/[\mathrm{AIBN}]_0$ (equiv.) & $T$ ($^\circ$C) & $t$ (h) & Conv. (\%)$^{a}$ & $M_n$ ($M_{n,\mathrm{theo}}$)$^{b,c}$ & $D^{b}$ \\
\midrule
1 & 2-(Methacryloyloxy)ethyl benzoate & 100/1 & 70 & 16 & 99 & 25{,}000 (23{,}200) & 2.48 \\
2 & 2-(Acryloyloxy)ethyl benzoate     & 100/1 & 70 & 16 & 99 & 16{,}200 (21{,}800) & 2.40 \\
3 & 2-Acrylamidoethyl benzoate       & 100/1 & 70 & 16 & 99 & 46{,}200 (21{,}700) & 3.33 \\
\bottomrule
\end{tabular}

\vspace{0.5em}
\footnotesize{
$^{a}$ Monomer conversions determined by $^1$H NMR. 
$^{b}$ PMMA-calibrated GPC values (DMF eluent). 
$^{c}$ Theoretical $M_n$ calculated according to $([\mathrm{DMDL}]_0/[\mathrm{CTA}]_0)\times(\text{monomer conversion})\times(\text{molecular weight of DMDL}) + (\text{molecular weight of CTA})$. \\
$D$ denotes dispersity ($M_w/M_n$).
}
\end{table}

\end{document}